\documentclass[11pt]{article}

\usepackage[preprint]{acl}

\usepackage{multirow}
\usepackage{xcolor,colortbl}
\definecolor{lavendergray}{rgb}{0.77, 0.76, 0.82}
\definecolor{lightgray}{rgb}{0.83, 0.83, 0.83}

\usepackage{array}
\newcolumntype{H}{>{\setbox0=\hbox\bgroup}c<{\egroup}@{}}
\newcommand{\tablestyle}[2]{\setlength{\tabcolsep}{#1}\renewcommand{\arraystretch}{#2}\centering\small}

\usepackage{amssymb}%
\usepackage{pifont}%
\usepackage{wrapfig}

\definecolor{myblue}{rgb}{0.11764705882352941, 0.5647058823529412, 1.0}
\definecolor{Gray}{gray}{0.9}
\definecolor{darkgreen}{rgb}{0.545, 0.749, 0.608}
\newcommand{\gain}[1]{{\color{darkgreen}\scriptsize\textbf{#1}}}
\newcommand{\loss}[1]{{\color{myblue}\scriptsize\textbf{#1}}}

\usepackage{tablefootnote}

\usepackage{soul}
\sethlcolor{Gray}
\usepackage{times}
\usepackage{latexsym}

\usepackage[T1]{fontenc}

\usepackage[utf8]{inputenc}

\usepackage{microtype}

\usepackage{inconsolata}

\usepackage{graphicx}
\usepackage{setspace}
\usepackage{subcaption}
\usepackage{hyperref}
\usepackage{url}
\usepackage{algorithm}
\usepackage{algpseudocode}
\usepackage{booktabs}      
\usepackage{multirow}      
\usepackage{multicol}      
\usepackage{colortbl}      
\usepackage{xcolor}        
\usepackage{graphicx}      
\usepackage{adjustbox}
\usepackage{array}      
\usepackage{booktabs}

\usepackage[utf8]{inputenc} 
\usepackage[T1]{fontenc}    
\definecolor{customblue}{rgb}{0.21,0.49,0.74}
\usepackage{hyperref}       
\usepackage{url}            
\usepackage{booktabs}       
\usepackage{amsfonts}       
\usepackage{nicefrac}       
\usepackage{microtype}      
\usepackage{graphicx}
\usepackage{amsmath}
\usepackage[hypcap=false]{caption}
\usepackage{nicematrix}

\title{Autoregressive Semantic Visual Reconstruction \\ Helps VLMs Understand Better}

\author{
    Dianyi Wang$^{1,2}$\thanks{Co-first authors.}\enskip\enskip 
    Wei Song$^{2,3,4}$\footnotemark[1]\enskip\enskip
    Yikun Wang$^{1,2}$\enskip\enskip
    Siyuan Wang$^{6}$\enskip\enskip\\
    {\bfseries Kaicheng Yu$^{3}$}\enskip\enskip
    {\bfseries Zhongyu Wei$^{1,2}$}\thanks{Corresponding author.}\enskip\enskip
    {\bfseries Jiaqi Wang$^{2,5}$}\footnotemark[2]\\
    \textsuperscript{1}Fudan University \enskip 
    \textsuperscript{2}Shanghai Innovation Institute \enskip
    \textsuperscript{3}AutoLab, Westlake University \enskip \\
    \textsuperscript{4}Zhejiang University \enskip
    \textsuperscript{5}Shanghai AI Lab \enskip
    \textsuperscript{6}University of Southern California \enskip \\
    \texttt{dywang24@m.fudan.edu.cn, songweii@zju.edu.cn}
}

\begin{document}
\maketitle
\begin{abstract}
Typical large vision-language models (LVLMs) apply autoregressive supervision primarily to textual responses, without fully exploiting causal learning over rich visual inputs. 
As a result, these models often emphasize vision-to-language alignment while potentially overlooking fine-grained visual information. While prior work has explored autoregressive image generation, effectively leveraging autoregressive visual supervision to enhance image understanding remains an open challenge. 
In this paper, we introduce Autoregressive Semantic Visual Reconstruction (ASVR), which enables joint learning of visual and textual modalities within a unified autoregressive framework. 
ASVR trains models to autoregressively reconstruct the semantic content of input images, which consistently enhances multimodal comprehension. Notably, we show that even when provided with continuous image features as input, models can effectively reconstruct discrete semantic tokens, resulting in stable and consistent improvements across various multimodal understanding benchmarks. ASVR delivers significant performance gains and scalability across varying data scales, visual input, visual supervision and model architectures. In particular, ASVR generally improves baselines by 2-3\% across 14 multimodal benchmarks.
\end{abstract}

\section{Introduction}
\label{sec:intro}


  

The success of large language models (LLMs) has demonstrated the tremendous potential and scalability of the autoregressive (AR) paradigm. Recent advances extending LLMs' powerful capabilities to multimodal understanding through bridge-style architectures, exemplified by LLaVA~\citep{liu2023llava,liu2024llava1d5,liu2024llavanext}, have achieved remarkable performance across vision-language tasks~\citep{liu2023mmbench,yue2023mmmu,fu2024mme,vqa_v2,POPE,gqa,kembhavi2016diagramworthdozenimages}. These models~\citep{Qwen-VL,Qwen2VL,yao2024minicpm,chen2024internvl,lu2024deepseekvl, wu2024deepseekvl2}, typically adopt a learnable projector to align features from a pretrained visual encoder into the text embedding space of LLMs.

However, most large vision-language models (LVLMs)\citep{wang2024qwen2vlenhancingvisionlanguagemodels,dong2024internlmxcomposer24khdpioneeringlargevisionlanguage,liu2024llava,li2024llavaonevisioneasyvisualtask} supervise only textual outputs through next-token prediction, while overlooking the causal learning of rich visual content within input images. Although this limitation is partially mitigated by training on image–caption pairs that associate visual content with language, the visual modality expresses far more than text alone, capturing spatial relationships, textures, complex compositions, and subtle stylistic cues that language cannot fully convey.
For example, LLaVA-1.5\cite{liu2023improved} represents a single 336$\times$336 image with 576 visual tokens, which collectively encode substantially more information than the associated caption, yet applies no explicit supervision to this visual content.

\begin{figure*}[t]
  \centering
  \includegraphics[width=1.0\linewidth]{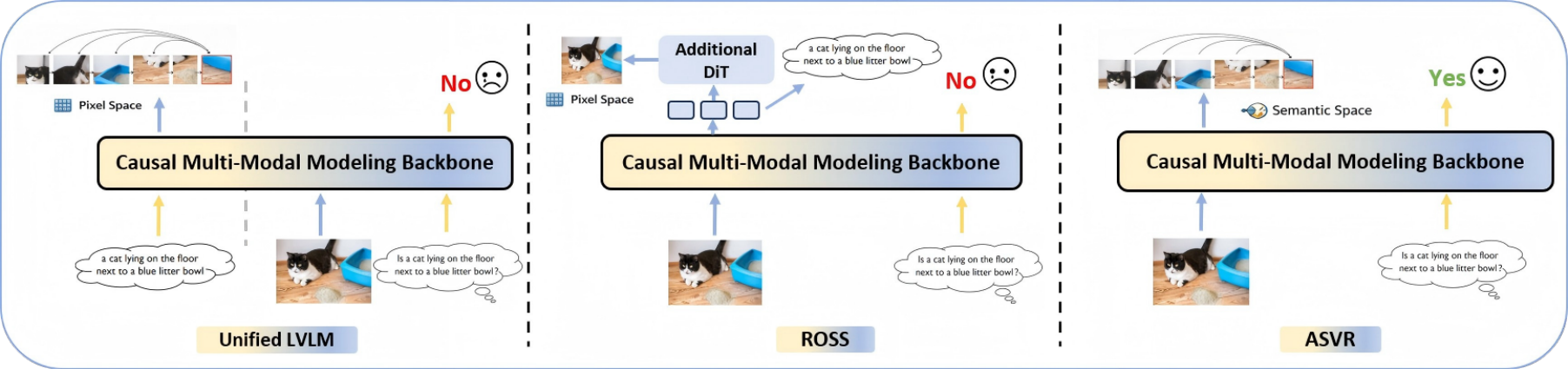}
  \caption{
    \textbf{(Left)} \textbf{Unified LVLMs} aim to implicitly improve visual understanding through additional text-to-image (T2I) objectives. \textbf{(Mid)} \textbf{ROSS} introduces an auxiliary diffusion transformer (DiT) to reconstruct pixel-level image information conditioned on latent representations from the LVLM. \textbf{(Right)} \textbf{ASVR} directly reconstructs the semantic content of the input image, enhancing semantic-level alignment.
  }
  \label{fig:teaser}
\end{figure*}


While recent unified models have explored integrating visual understanding and generation within the autoregressive paradigm~\citep{team2024chameleon, wang2024emu3, wu2024vilau, metamorph}, visual tokens are typically supervised only on the output side through visual generation objectives. In contrast, visual tokens fed as inputs are not explicitly supervised to enhance visual understanding.
Effectively supervising autoregressive visual inputs to improve fine-grained visual understanding remains an open challenge. Most recently, \citet{wang2024ross} proposed reconstructing visual inputs via denoising, yet their method relies on external Diffusion Transformer (DiT) modules and lacks a unified next-token prediction framework that explicitly encourage causal learning over detailed visual information and dependencies rather than high-level bidirectional contextual correlations.

In this paper, we introduce \textbf{Autoregressive Semantic Visual Reconstruction (ASVR)}, a method that enables joint learning of the visual input and textual output within a unified autoregressive paradigm, without relying on any external modules. Specifically, ASVR allows LVLMs to autoregressively predict the next discrete semantic token of an input image, which is prepared by a pretrained semantic visual tokenizer~\citep{song2025dualtoken,wu2024vilau,qu2024tokenflow,xie2024musevl}.
Autoregressively reconstructing semantic visual representations consistently enhances the visual understanding capabilities and further improve reasoning performance. Notably, we find that models can effectively reconstruct discrete semantic tokens even when provided with continuous image features as input. This setting yields substantial gains over approaches that use shared discrete semantic visual tokens for both input and output, and also outperforms the denoising-based visual reconstruction paradigm~\citep{wang2024reconstructivevisualinstructiontuning}.

Our approach delivers significant and consistent gains across varying data scales (LLaVA-1.5-665K~\citep{liu2023improved}, LLaVA-Next-779K~\citep{liu2024llava}, Bunny-v1\_1-data-2M~\citep{he2024efficientmultimodallearningdatacentric},LLaVA-OV-4M~\citep{li2024llavaonevisioneasyvisualtask}) as well as across model architectures including Vicuna family~\citep{zheng2023judgingllmasajudgemtbenchchatbot} and Mistral~\citep{jiang2023mistral}. Specifically, \textbf{ASVR} improves baselines by 2-3\% on average across 14 multimodal benchmarks  with improvements remaining robust across visual feature types, LLM backbone capacities, data scales, and high-resolution scenarios.
These results underscore the importance of explicit semantic visual supervision in training LVLMs. ASVR not only improves visual understanding but also introduces a scalable, unified training strategy, offering a new perspective on autoregressive modeling for multimodal systems.

\section{Preliminaries}
\label{Preliminaries}
\paragraph{Large Vision Language Models}
\label{ Large Vision Language Models Modeling}
To process and represent input from different modalities in a unified manner, LVLMs typically comprise three components: a pre-trained LLM core, a projector commonly implemented as a two-layer MLP, and a pre-trained visual encoder with semantic alignment.

Given a input RGB image $I \in \mathbb{R}^{H \times W \times 3}$, where $H$ and $W$ denote the image height and width, a pre-trained visual encoder $V_{\xi}$ first  extracts visual features $\mathbf{z}^I = V_{\xi}(I)$. These features are then projected into the LLM embedding space through a projector $P_{\phi}$, yielding a sequence of visual embeddings: $\mathbf{H}^I = P_{\phi}(\mathbf{z}^I) \in \mathbb{R}^{ m \times d}$, where $m = h \times w$ denotes the length of visual features, and $d$ is the embedding dimension of LLM. $\xi$ and $\phi$ are the parameters of the visual encoder and projector, respectively. For a textual input $T \in \mathbb{Z}^{L}$, the LLM tokenizer produces a sequence of token indices $\mathbf{x}^T = \text{Tokenizer}(T) \in \mathbb{R}^n$. which are then mapped into textual embeddings via the LLM’s embedding layer $\mathbf{H}^T = \text{Embedding}({x}^T) \in \mathbb{R}^{n \times d}$ where $n$ denotes the text sequence length.

The multimodal input is formed by concatenating the visual and textual embeddings as $[\mathbf{H}^I, \mathbf{H}^T] \in \mathbb{R}^{(m+n) \times d}$, which is then fed into a causal LLM backbone $L_{\theta}$ with parameters $\theta$ for unified autoregressive modeling:
\begin{equation}
L_{\theta}([\mathbf{H}^I, \mathbf{H}^T]) = \prod_{i=1}^{n} L_{\theta}(x_i^T \mid x_{<i}^T, \mathbf{H}^I)
\end{equation}

\paragraph{Training Pipeline for LVLMs}
LVLMs training typically follows a two-stage paradigm~\citep{liu2023llava}: pre-training and instruction tuning. During pre-training, the model learns to align visual and textual modalities, enabling joint understanding of multimodal inputs. Instruction tuning further enhances generalization across diverse downstream tasks such as visual question answering (VQA). The training objective is to maximize the likelihood of target textual responses in an autoregressive manner, with supervision applied only to textual responses.
In practice, pre-training usually updates only the projector parameters $\phi$, while instruction tuning additionally fine-tunes the LLM parameters $\theta$. The visual encoder $v_{\xi}$ may either remain frozen~\citep{liu2023llava,tong2024cambrian1fullyopenvisioncentric} or be jointly optimized~\citep{li2024llavaonevisioneasyvisualtask,dong2024internlmxcomposer24khdpioneeringlargevisionlanguage,wang2024qwen2vlenhancingvisionlanguagemodels,liu2024llava}. 

\section{Method}
\label{method}
In this section, we introduce \textbf{ASVR} which learns autoregressive modeling of textual responses while simultaneously reconstructs visual inputs autoregressively  to enhance visual understanding. An overview of the method is provided in Section~\ref{overview}, followed by detailed descriptions of the visual tokenizer and visual encoder in Sections~\ref{tokenizer} and~\ref{encoder}, respectively. The training procedure is described in Section~\ref{training recipe}. Figure~\ref{fig:arch} provides a detailed comparison between conventional LVLMs (e.g., LLaVA) and our ASVR, highlighting the key innovation of incorporating autoregressive visual input supervision to enhance multimodal understanding.

\subsection{Overview}
\label{overview}
\begin{figure*}[t]
    \centering
    \includegraphics[width=0.92\linewidth]{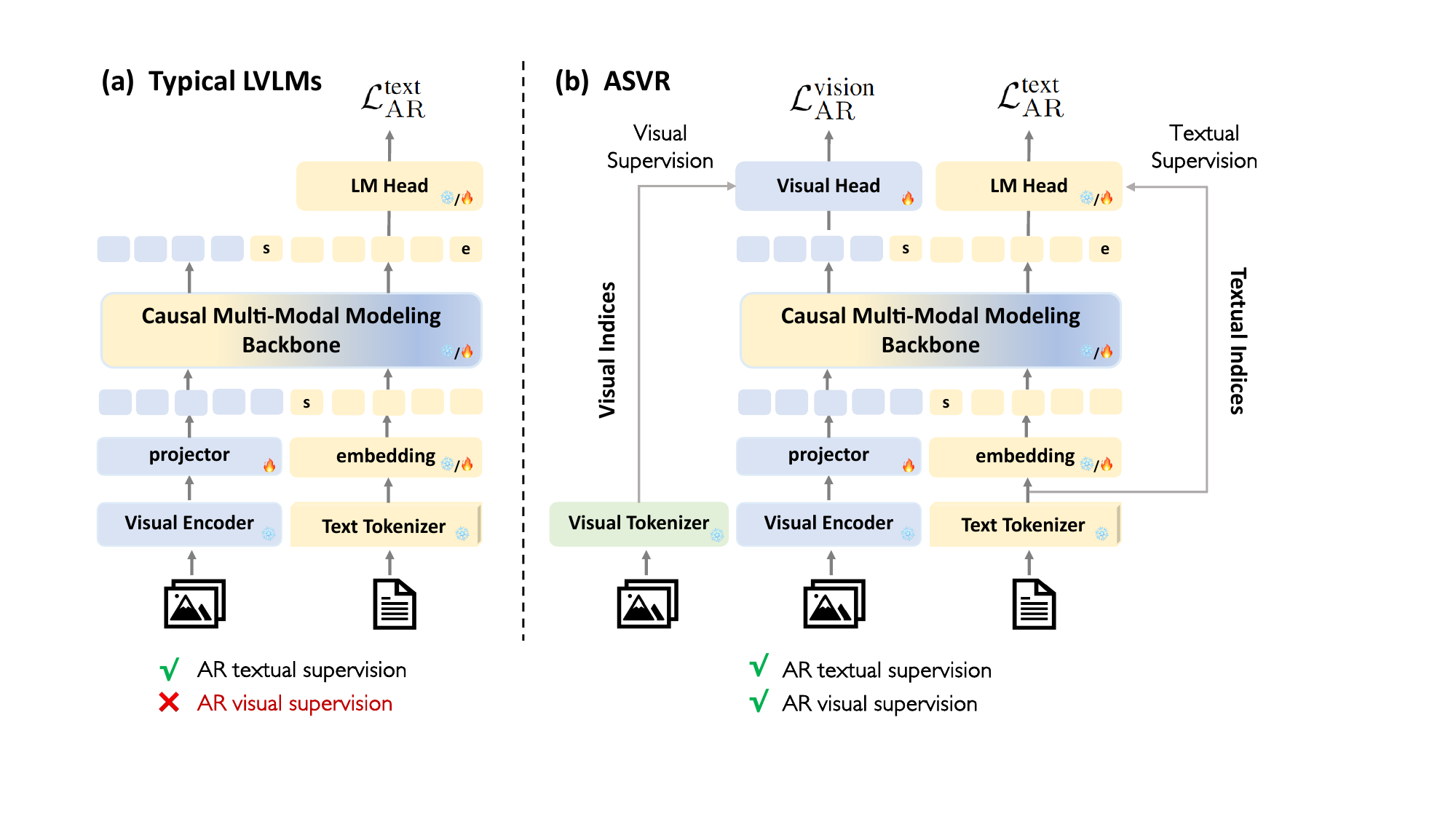} 
    \caption{\textbf{Left}: The typical LVLM paradigm such as LLaVA~\citep{liu2023llava}. \textbf{Right}: Overview of \textbf{ASVR's}  architecture and training pipeline. The input image and text are tokenized into sequences of discrete token indices enabling unified autoregressive supervision over both visual inputs and textual outputs. For each module, the icon before the slash indicates whether it is frozen or tunable during pre-training, while the icon after the slash indicates its configuration during instruction tuning. "s" and "e" denote the start and end of the text tokens, respectively.}
    \label{fig:arch}
\end{figure*}
We incorporate autoregressive visual supervision into typical LVLMs described in Section~\ref{Preliminaries} by extending the next-token prediction paradigm to reconstruct visual inputs. This unified formulation enables the model to seamlessly integrate visual and textual information, thereby establishing a perceptual foundation for image understanding, alleviating information loss caused by text-only supervision, and ultimately enhancing multimodal understanding  and reasoning capabilities.

As illustrated in Figure~\ref{fig:arch}(b), we employ a visual tokenizer to convert the input image $I$ into a discrete sequence of visual token indices $\mathbf{x}^I \in \mathbb{R}^m$, which serve as visual supervision signals and $m$ matches the length of the visual features sequence $\mathbf{H}^I$ extracted by the pre-trained visual encoder and fed into the LLM backbone. A visual prediction head tailored to the visual tokenizer is trained to autoregressively predict the next visual token, analogous to textual supervision:
\begin{equation}
\mathcal{L}_{\mathrm{AR}}^{\text{vision}}(\Theta, I) = - \frac{1}{m} \sum_{i=1}^{m} \log L_{{\theta}}(x_i^I \mid x_{<i}^I),
\end{equation}
where $\Theta = \{\theta, \xi, \phi\}$ denotes the parameters of the LLM backbone, visual encoder and projector. The final training objective jointly optimizes visual and textual autoregressive losses, $\mathcal{L}_{\mathrm{AR}}^{\text{vision}}$ and $\mathcal{L}_{\mathrm{AR}}^{\text{text}}$:
\begin{equation}
\mathcal{L}_{\mathrm{AR}}(\Theta, I,T) = \mathcal{L}_{\mathrm{AR}}^{\text{vision}} + \mathcal{L}_{\mathrm{AR}}^{\text{text}}
\end{equation}
This design unifies the learning paradigm across modalities, and encourages the model to first develop a coherent visual understanding, which subsequently serves as a foundation for more accurate and contextually grounded multimoda reasoning.

\subsection{Visual Tokenizer}
\label{tokenizer}
Unlike continuous visual encoders, a visual tokenizer converts an input image into a one-dimensional sequence of discrete visual codes through vector quantization (VQ) by learning a fixed-size visual codebook. The corresponding embeddings are retrieved from the codebook based on these codes and used as inputs to the LLM backbone. In this work, we utilize such a visual tokenizer to obtain discrete visual token representations
as supervision targets for visual reconstruction. Existing visual tokenizers can be broadly categorized into two types.

\paragraph{Visual Appearance Tokenizer} A visual appearance tokenizer~\citep{esser2021vqgan,team2024chameleon} is optimized with the objective of reconstructing the appearance of the input image, typically using a combination of pixel-wise L2 loss~\citep{dosovitskiy2016L2}, LPIPS loss\cite{zhang2018lpips} and adversarial loss \cite{isola2017gan}. The resulting sequence of token indices corresponds to a quantized mapping of the image’s pixel-level features. Using appearance tokenizers to provide pixel-level supervision will encourage LVLMs to focus on low-level visual feature reconstruction.

\paragraph{Visual Semantic Tokenizer} A visual semantic tokenizer~\citep{qu2024tokenflow,wu2024vilau,xie2024musevl,song2025dualtoken} is trained to align image features with textual semantics, typically using a contrastive loss~\citep{radford2021clip} to enhance cross-modal alignment. The resulting sequence of token indices represents a quantized mapping of the image’s high-level semantic features. Using semantic tokenizers to provide semantic visual supervision will guide LVLMs to focus on semantically meaningful structures and relationships of the image, thereby promoting more effective multimodal understanding. 


\subsection{Visual Encoder}
\label{encoder}
The visual encoder provides continuous visual features as inputs to the LLM backbone, directly affecting the quality of visual modeling. To enhance multimodal understanding, it is crucial to employ a visual encoder that is semantically aligned with textual representations~\citep{wu2024vilau,qu2024tokenflow,wu2024janus}, allowing the extraction of high-level and semantically meaningful image features. Typically, such visual encoders adopt transformer-based~\citep{dosovitskiy2021imageworth16x16words} architecture, trained with contrastive loss~\citep{radford2021clip} to align closely with textual semantics and directly convert input images into one-dimensional sequences of continuous feature vectors.



\subsection{Training Recipe}
\label{training recipe}
Our training recipe is illustrated in Figure~\ref{fig:arch}, which extends the standard LVLM training by incorporating visual supervision to enable unified autoregressive modeling over both visual inputs and textual responses. Specifically, during the pre-training stage, we focus solely on optimizing the projector and the visual prediction head. This stage aligns visual representations sequence with the LVLM’s semantic space, allowing the model to acquire an initial visual perception by learning the mapping from continuous visual features to discrete visual token indices.
In the instruction tuning stage, we further fine-tune the LLM backbone parameters. By leveraging diverse vision-language instruction data, the model is encouraged to perform deeper semantic understanding of visual content, thereby enhancing its ability to understand and reason across modalities in a more comprehensive manner.

\begin{table*}[th!]
\centering
\caption{\textbf{Impact of ASVR with different combinations of visual encoders and visual tokenizers across multimoda understanding benchmarks.}
``\ding{55}'' indicates training with textual supervision only, while ``\ding{51}'' denotes the inclusion of visual supervision via an additional $\mathcal{L}_{\mathrm{AR}}^{\text{vision}}$.
``Sem.'' refers to using visual semantic tokenizer to construct visual supervision targets; "App." denotes a visual appearance tokenizer;
"App.+Sem." indicates dual supervision, where visual semantic and visual appearance tokenizers are used independently to compute their respective $\mathcal{L}_{\mathrm{AR}}^{\text{vision}}$, which are then summed. Our proposed ASVR utilize semantic supervision.}
\tablestyle{5pt}{1.5}
\setlength\tabcolsep{2pt}
\resizebox{1\textwidth}{!}{
\begin{tabular}{ccc|llll|llll|ll|ll|ll|c}
\toprule

{\multirow{-2}{*}{}}                         & {} & {}                 & \multicolumn{4}{c}{OCR}                                                  & \multicolumn{4}{c}{General}                                        & \multicolumn{2}{c}{Knowledge}            & \multicolumn{2}{c}{Visual-Centric}                & \multicolumn{2}{c|}{Hallusion}                      &    \\ \cline{4-17}

{}                                          & {\multirow{-2}{*}{$\mathcal{L}_{\mathrm{AR}}^{\text{vision}}$}}                                 & {\multirow{-2}{*}{\begin{tabular}[c]{@{}c@{}}Visual \\ Tokenizer\end{tabular}}}                                                & TVQA       & DVQA        & OCRB      & {CQA}       & MMB       & MME           & SEED    & {GQA}           & MMMU     & {AI2D}          & RQA   & {MMVP}          & Hbench & {POPE}          & \multirow{-2}{*}{AVG}   \\

\midrule
\multicolumn{18}{c}{\cellcolor[HTML]{DEE0E3}Visual Encoder: VQ-SigLIP-ViT-SO400M/14@384  (Discrete Visual Features)}                                                                                                                                                                                                                                                                                                                                                                                                                                                                               \\ \midrule
{\cellcolor[HTML]{FFFFFF}LLaVA}         & {\cellcolor[HTML]{FFFFFF}\ding{55}}        & {\cellcolor[HTML]{FFFFFF}-}                        & 49.3          &20.0               &29.5               & {12.4}              & 60.4          & 56.9          & 63.1          & {56.2}          & 31.2          & {50.4}          &50.2               & {24.7}              & 21.8               & {80.7}          &43.3  \cellcolor[HTML]{FFFFFF} \\

\rowcolor[HTML]{E1EAFF} 
{\textbf{ASVR}} & {\ding{51}}        & {Sem.} & 55.5\gain{(+6.2)}          &21.4\gain{(+1.4)}               &32.4\gain{(+2.9)}               & {14.7}\gain{(+2.3)}               & 62.3\gain{(+1.9)}           & 57.7\gain{(+0.8)}           & 65.4\gain{(+2.3)}          & {57.1}\gain{(+0.9)}           & 32.0\gain{(+0.8)}          & {53.5}\gain{(+3.1)}           &52.3\gain{(+2.1)}               & {26.0}\gain{(+1.3)}              &27.7\gain{(+5.9)}                & {76.8}\loss{(-3.9)}          &45.3  \\ \midrule
\multicolumn{18}{c}{\cellcolor[HTML]{DEE0E3}Visual Encoder: SigLIP-ViT-SO400M/14@384 (Continuous Visual Features)}                                                                                                                                                                                                                                                                                                                                                                                                                                                              \\ \midrule
{LLaVA}                                 & {\ding{55}}                                & {-}                                                & 56.0          & 21.1          & 31.3          & {14.6}          & 64.0          & 67.2          & 63.8          & {60.5}          & 32.7          & {53.5}          & 52.0          & {28.7}          & 23.9           & {85.9}          & 46.8                     \\
{Appearance Supervise}                            & {\ding{51}}                                & {App.}                            & 53.7\loss{(-2.3)}            & 17.8\loss{(-3.3)}          & 30.2\loss{(-1.1)}          & {14.4}\loss{(-0.2)}          & 61.6\loss{(-2.4)}          & 68.7\loss{(-1.5)}          & 59.5\loss{(-4.3)}          & {57.8}\loss{(-2.7)}          & 33.1\gain{(+0.4)}          & {53.7}\gain{(+0.2)}          & 49.3\loss{(-2.7)}          & {22.0}\loss{(-6.7)}          & 24.0\gain{(+0.1)}           & {84.1}\loss{(-1.8)}          & 45.0                     \\
{Dual Supervise}                             & {\ding{51}}                                & {App.+Sem.}           & 59.4\gain{(+3.4)}          & 23.7\gain{(+2.6)}          & 33.5\gain{(+2.2)}          & {16.1}\gain{(+1.5)}          & 65.6\gain{(+1.6)}          & 70.2\gain{(+3.0)}          & 66.1\gain{(+2.3)}          & {61.5}\gain{(+1.0)}          & \textbf{34.0}\gain{(+1.3)} & {56.3}\gain{(+2.8)}          & 53.5\gain{(+1.5)}          & {22.0}\loss{(-6.7)}          & 30.7\gain{(+6.8)}           & {86.3}\gain{(+0.4)}          & 48.5                     \\
\rowcolor[HTML]{E1EAFF} 
{\textbf{ASVR}}                         & {\ding{51}}                                & {Sem.}                         & \textbf{59.5}\gain{(+3.5)}  & \textbf{24.3}\gain{(+3.2)} & \textbf{35.4}\gain{(+4.1)} & {\textbf{16.4}}\gain{(+1.8)} & \textbf{66.1}\gain{(+2.1)} & \textbf{72.8}\gain{(+5.6)} & \textbf{66.4}\gain{(+2.6)} & {\textbf{61.5}}\gain{(+1.0)} & 33.9\gain{(+1.2)}          & {\textbf{57.0}}\gain{(+3.5)} & \textbf{54.1}\gain{(+2.1)} & {\textbf{30.0}}\gain{(+1.3)} & \textbf{33.7}\gain{(+9.8)}  & {\textbf{86.3}}\gain{(+0.4)} & \textbf{49.8}            \\ \midrule
\end{tabular}}
\label{table: main results}
\end{table*}
\section{Experiments}
\label{experiment}

In this section, we present a comprehensive set of controlled experiments to evaluate the effectiveness of our method~\textbf{(ASVR)} within typical LVLM's frameworks~\citep{liu2023llava} across a diverse range of multimoda understanding tasks. We begin by detailing our experimental setup. Then, we analyze the impact of different visual encoders and visual tokenizers on the model's performance. Finally, we further validate the generalization and adaptability of our method across various LLM backbones with different parameter scales and under varying amounts of training data.

\subsection{Experimental Setup}
\label{subsec:setup}
\paragraph{Implementation Details.} We implement our experiments baseline on the LLaVA-1.5~\citep{liu2023improved} with only textual supervision as discussed in sec~\ref{Preliminaries}. We utilize vicuna-v1.5-7B \citep{zheng2023judgingllmasajudgemtbenchchatbot} as the LLM backbone and initialize visual encoder with the pretrained weights from SigLIP-SO400M-patch14-384 \citep{alabdulmohsin2023so400m} to support continuous visual features for LVLMs. For visual tokenizer, we employ both visual appearance tokenizer and visual semantic tokenizer(VQ-SigLIP)proposed in DualToken~\citep{song2025dualtoken} to construct visual supervision targets, which convert input images into $27 \times 27 \times 8$ visual semantic or appearance token sequences, with a residual depth of $D = 8$. The visual head also derived from DualToken, is integrated and aligned with the chosen visual tokenizer to ensure architectural compatibility. Training is conducted on LLaVA-558K~\citep{liu2023llava} for pretraining and LLaVA-1.5-665K~\citep{liu2023llava} for instruction tuning. 

\paragraph{Evaluation Details}
We conduct a comprehensive evaluation of model's capabilities on 14 widely used vision-language understanding benchmarks. Specifically, the general multimodal benchmarks include MMBench~\citep{liu2024mmbenchmultimodalmodelallaround} English dev split(MMB), GQA~\citep{hudson2019gqanewdatasetrealworld}, SEED-Image(SEED)~\citep{li2023seedbenchbenchmarkingmultimodalllms} and MME sum~\citep{fu2024mmecomprehensiveevaluationbenchmark}. For OCR-based question answering, we assessed performance on TextVQA(TVQA)~\citep{singh2019vqamodelsread}, ChartQA(CQA)~\citep{masry2022chartqabenchmarkquestionanswering},  DocVQA(DVQA)~\citep{mathew2021docvqadatasetvqadocument} and OCRBench(OCRB)~\citep{Liu_2024} . For knowledge-based question answering, we utilize MMMU validation split~\citep{yue2024mmmumassivemultidisciplinemultimodal}, AI2D~\citep{kembhavi2016diagramworthdozenimages}. Additionally, we evaluated hallucination robustness on POPE~\citep{li2023evaluatingobjecthallucinationlarge}, Hallusionbench(Hbench)~\citep{guan2024hallusionbenchadvanceddiagnosticsuite} and visual-centric tasks on MMVP~\citep{tong2024eyeswideshutexploring} and RealworldQA(RQA)~\citep{xai2024grok}. Evaluation prompts can be found in Appendix~\ref{eval_prompts}.

\begin{table*}[th!]
\centering
\caption{\textbf{Generalizability of ASVR to different training data scales and LLM backbones across benchmarks.}
The same visual encoder (SigLIP-ViT-SO400M/14@384) is used for ASVR and the baseline. ``/'' separates data scales for pre-training (left), mid-training (middle, if applicable) and instruction tuning (right).}
\tablestyle{5pt}{1.5}
\setlength\tabcolsep{2pt}
\resizebox{1\textwidth}{!}{
\begin{tabular}{cccc|llll|llll|ll|ll|ll|c}
\toprule
{}                                           & {}                                 & {}                                                                          & {}                                                                        & \multicolumn{4}{c}{OCR}                                   & \multicolumn{4}{c}{General}                          & \multicolumn{2}{c}{Knowledge}              & \multicolumn{2}{c}{Visual-centric}     & \multicolumn{2}{c|}{Hallusion}             &                              \\ \cline{5-18}
{\multirow{-2}{*}{}}                         & {\multirow{-2}{*}{$\mathcal{L}_{\mathrm{AR}}^{\text{vision}}$}} & {\multirow{-2}{*}{\begin{tabular}[c]{@{}c@{}}LLM \\ backbone\end{tabular}}} & {\multirow{-2}{*}{\begin{tabular}[c]{@{}c@{}}Data \\ Scale\end{tabular}}} & TVQA & DVQA & OCRB & {CQA} & MMB & MME  & SEED & {GQA}  & MMMU & {AI2D} & RQA & {MMVP} & Hbench & {POPE} & \multirow{-2}{*}{AVG}        \\ \midrule
\multicolumn{19}{c}{\cellcolor[HTML]{DEE0E3}With Different Data Scales}                                                                                                                                                                                                                                                                                                                                                                                                                                                                                                                                           \\ \midrule
{\cellcolor[HTML]{FFFFFF}Baseline}         & {\cellcolor[HTML]{FFFFFF}\ding{55}}        & {\cellcolor[HTML]{FFFFFF}vicuna-v1.5-7B}                                    & {\cellcolor[HTML]{FFFFFF}2M/2M}                                              & \textbf{61.6}    & \textbf{43.8}   & 35.4     & {38.7}    & 68.4    & 74.9 & 67.9     & {61.7} & 40.6      & {64.6}       & \textbf{56.1}        & {34.8} & 36.9           & {85.6} & \cellcolor[HTML]{FFFFFF}55.1 \\ 

\rowcolor[HTML]{E1EAFF}
{\textbf{ASVR}} & {\ding{51}}        & {vicuna-v1.5-7B}                                    & {2M/2M}                                              & 60.6\loss{(-1.0)}    & 43.1\loss{(-0.7)}   & \textbf{36.2}\gain{(+0.8)}     & \textbf{38.9}\gain{(+0.2)}    & \textbf{68.6}\gain{(+0.2)}     & \textbf{76.2}\gain{(+1.3)} & \textbf{68.7}\gain{(+0.8)}     & \textbf{62.0}\gain{(+0.3)} & \textbf{41.4}\gain{(+0.8)}      & \textbf{64.8}\gain{(+0.2)}       & 55.9\loss{(-0.2)}        & \textbf{35.9}\gain{(+1.1)} & \textbf{42.2}\gain{(+5.3)}           & \textbf{85.7}\gain{(+0.1)} & \textbf{55.7} \\ \midrule
Baseline             & \ding{55}                               & vicuna-v1.5-7B               & 558K/4M/4M                        & 57.2          & 44.1          & 49.6          & \multicolumn{1}{l|}{39.2}          & 71.7         & 71.7          & 68.7          & \multicolumn{1}{l|}{58.2}          & 37.9          & \multicolumn{1}{l|}{70.7}        & 56.7        & \multicolumn{1}{l|}{40.0}            & 36.1           & 85.5          & 56.2                                    \\
\rowcolor[HTML]{E1EAFF} 
\textbf{ASVR}              & \ding{51}                               & vicuna-v1.5-7B               & 558K/4M/4M                        & \textbf{60.0\gain{(+2.8)}} & \textbf{46.5\gain{(+2.4)}} & \textbf{51.3\gain{(+1.7)}} & \multicolumn{1}{l|}{\textbf{41.7\gain{(+2.5)}}} & \textbf{72.2\gain{(+0.5)}} & \textbf{73.2\gain{(+1.5)}} & \textbf{69.9\gain{(+1.2)}} & \multicolumn{1}{l|}{\textbf{59.8\gain{(+1.6)}}} & \textbf{39.7\gain{(+1.8)}} & \multicolumn{1}{l|}{\textbf{71.8\gain{(+1.1)}}} & \textbf{57.5\gain{(+0.8)}} & \multicolumn{1}{l|}{\textbf{42.0}\gain{(+2.0)}} & \textbf{37.9}\gain{(+1.8)}  & \textbf{86.9}\gain{(+1.4)} & \textbf{57.9}                            \\ \midrule
\multicolumn{19}{c}{\cellcolor[HTML]{DEE0E3}With Different LLM Backbones}                                                                                                                                                                                                                                                                                                                                                                                                                                                                                                                                         \\ \midrule
{Baseline}                                 & {\ding{55}}                                & {Mistral-7B}                                                                & {558K/665K}                                                                    & 50.8    & 15.7   & \textbf{34.6}     & {15.2}    & 65.9    & 66.9 & 67.9     & {62.4} & 32.0      & {53.0}       & 55.0        & {35.3} & 32.7           & {86.6} & 48.1                         \\ 

\rowcolor[HTML]{E1EAFF}
{\textbf{ASVR}}                         & {\ding{51}}                                & {Mistral-7B}                                                                & {558K/665k}                                                                    & \textbf{54.9}\gain{(+4.1)}    & \textbf{17.9}\gain{(+2.2)}   & 34.1\loss{(-0.5)}     & \textbf{15.6}\gain{(+0.4)}    & \textbf{67.1}\gain{(+1.2)}    & \textbf{71.5}\gain{(+4.6)} & \textbf{68.3}\gain{(+0.4)}     & \textbf{62.5}\gain{(+0.1)} & \textbf{32.6}\gain{(+0.6)}      & \textbf{54.5}\gain{(+1.5)}       & \textbf{55.4}\gain{(+0.4)}        & \textbf{35.7}\gain{(+0.4)} & \textbf{35.0}\gain{(+2.3)}           & \textbf{86.8}\gain{(+0.2)} & \textbf{49.4}                         \\ \midrule
Baseline             & \ding{55}                               & Vicuna-v1.5-13B               & 558K/665k                        & 57.2          & 22.1          & 32.4          & \multicolumn{1}{l|}{15.1}          & 67.1          & 68.9          & 65.6          & \multicolumn{1}{l|}{60.4}          & 35.6          & \multicolumn{1}{l|}{54.9}        & 54.8        & \multicolumn{1}{l|}{34.0}            & 32.9           & 86.8          & 49.1                                     \\
\rowcolor[HTML]{E1EAFF} 
\textbf{ASVR}              & \ding{51}                               & Vicuna-v1.5-13B               & 558k/665K                        & \textbf{61.6\gain{(+4.4)}} & \textbf{27.3\gain{(+5.2)}} & \textbf{37.1\gain{(+4.7)}} & \multicolumn{1}{l|}{\textbf{18.4\gain{(+3.3)}}} & \textbf{70.8\gain{(+3.7)}} & \textbf{74.9\gain{(+6.0)}} & \textbf{68.7\gain{(+3.1)}} & \multicolumn{1}{l|}{\textbf{62.8\gain{(+2.4)}}} & \textbf{36.4\gain{(+0.8)}} & \multicolumn{1}{l|}{\textbf{60.0\gain{(+5.1)}}} & \textbf{56.0\gain{(+1.2)}} & \multicolumn{1}{l|}{\textbf{35.3}\gain{(+1.3)}} & \textbf{36.8}\gain{(+3.9)}  & \textbf{87.5}\gain{(+0.7)} & \textbf{52.4}                            \\ \midrule
\end{tabular}}
\label{table:Generality}
\end{table*}
\subsection{Main Results}
\paragraph{The Effectiveness of ASVR}
As shown in Table~\ref{table: main results}, with the configuration of the continuous-based visual encoder (SigLIP), we observe ASVR consistent and significant performance improvements across all 14 benchmarks, increasing the average score from \textbf{46.8} to \textbf{49.8}, with 3\%. Notably, the gains are evident even on knowledge-based QA such as MMMU~\citep{yue2024mmmumassivemultidisciplinemultimodal} and AI2D~\citep{kembhavi2016diagramworthdozenimages}, suggesting that reconstructing and  and perceiving visual inputs can enhance the model’s cognitive reasoning abilities. Furthermore, substantial improvements are also observed on fine-grained tasks such as OCRBench~\citep{Liu_2024}, MMVP~\citep{tong2024eyeswideshutexploring}, and HallusionBench~\citep{guan2024hallusionbenchadvanceddiagnosticsuite}. In particular, HallusionBench sees an increase of nearly 10 points, further validating the effectiveness of our method. Moreover, under the configuration with a discrete-based visual encoder(VQ-SigLIP), semantic visual supervision also yields notable performance gains over the baseline. This further demonstrates the generalizability and robustness of our method.

\paragraph{Semantic v.s. Appearance} Specifically, ASVR incorporating semantic supervision alone yields the highest average performance across benchmarks, outperforming even the dual supervision setting that combines both appearance and semantic visual indices. In contrast, applying appearance-only supervision degrades model performance compared to the baseline. These results highlight that guiding the LVLM to reconstruct and perceive high-level semantic visual information of the input image, rather than low-level appearance details, more effectively enhances its multimoda understanding capabilities.

\paragraph{Continuous vs. Discrete}
We adopt SigLIP-ViT-SO400M/14@384~\citep{zhai2023siglip} to provide continuous visual features, while employing visual semantic tokenizer VQ-SigLIP~\citep{song2025dualtoken} to generate discrete visual features; both approaches aligned with textual semantics. Our experimental results indicate that, regardless of whether autoregressive semantic visual supervision is applied, the configuration of using continuous visual features consistently outperforms its discrete features counterpart arcoss all benchmarks. This performance gap may be attributed to image feature degradation introduced by vector quantization in discrete encoding, which can lead to loss of fine-grained visual information crucial for downstream multimoda understanding. More ablations on semantic tokenizers, training strategies and visual supervision are provided in Appendix~\ref{ablation}.

\paragraph{Discussion} The combination of visual encoder for provide visual features and visual semantic tokenizer for constructing semantic visual supervision targets proves to the most effective model configuration. The visual encoder avoids the visual information loss typically introduced by vector quantization, thereby providing better visual inputs for the LMM. Meanwhile, semantic supervision guides the LVLM reconstruct high-level, semantically meaningful aspects of the image, which are benefit for multimoda understanding.Notably, our findings demonstrate that continuous visual inputs with discrete semantic visual supervision targets can be seamlessly integrated into the unified autoregressive next-token prediction paradigm in the same manner as language. This formulation enables the LVLM to reconstruct and perceive visual semantic information, enhancing LVLM's capacity for comprehensive multimoda understanding. We further demonstrate that the unified autoregressive modeling paradigm consistently surpasses its denoising-based counterpart~\citep{wang2024reconstructivevisualinstructiontuning}, with results provided in the Section~\ref{app:ross}.

\begin{table*}[th!]
\centering
\caption{\textbf{High resolution adaptation of ASVR across multimoda understanding benchmarks.} We follow LLaVA-Next~\cite{liu2024llava} that utilize the visual encoder(SigLIP-ViT-SO400M/14@384) and high resolution input (1152 × 1152) for ASVR and baseline.}
\tablestyle{5pt}{1.5}
\setlength\tabcolsep{2.2pt}
\resizebox{1\textwidth}{!}{
\begin{tabular}{cccc|llllllllllllll|l}
\toprule
\multirow{2}{*}{}  & \multirow{2}{*}{$\mathcal{L}_{\mathrm{AR}}^{\text{vision}}$} & \multirow{2}{*}{LLM backbone} & \multirow{2}{*}{Data Scale} & \multicolumn{4}{c}{OCR}                                                                                       & \multicolumn{4}{c}{General}                                                                       & \multicolumn{2}{c}{Knowledge}                          & \multicolumn{2}{c}{Visual-centric}                          & \multicolumn{2}{c|}{Hallusion}                                 & \multicolumn{1}{c}{\multirow{2}{*}{AVG}} \\ \cline{5-18}
                   &                                 &                               &                             & \multicolumn{1}{l}{TVQA} & \multicolumn{1}{l}{DVQA} & \multicolumn{1}{l}{OCRB} & \multicolumn{1}{l|}{CQA} & \multicolumn{1}{l}{MMB} & \multicolumn{1}{l}{MME} & \multicolumn{1}{l}{SEED} & \multicolumn{1}{l|}{GQA}  & \multicolumn{1}{l}{MMMU} & \multicolumn{1}{l|}{AI2D} & \multicolumn{1}{l}{RQA} & \multicolumn{1}{l|}{MMVP} & \multicolumn{1}{l}{Hbench} & \multicolumn{1}{l|}{POPE} & \multicolumn{1}{c}{}                     \\ \midrule
LLaVA              & {\ding{55}}                               & Vicuna-v1.5-7B                & 558K/779k                        & 58.1                        & 44.1                       & 39.5                         & \multicolumn{1}{l|}{47.5}    & 66.6                        & 74.1                    & 66.8                         & \multicolumn{1}{l|}{62.0} & 35.8                          & \multicolumn{1}{l|}{62.8}       & \textbf{57.8}                            & \multicolumn{1}{l|}{30.0} & 40.6                               & 84.5                      & 55.0                                     \\

\rowcolor[HTML]{E1EAFF}
\textbf{ASVR} & {\ding{51}}                               & Vicuna-v1.5-7B                & 558K/779k                        & \textbf{58.9}\gain{(+0.8)}                        & \textbf{48.9}\gain{(+4.8)}                       & \textbf{45.6}\gain{(+6.1)}                         & \multicolumn{1}{l|}{\textbf{49.3}\gain{(+1.8)} }    & \textbf{68.0}\gain{(+1.4)}                        & \textbf{76.7}\gain{(+2.6)}                    & \textbf{67.2}\gain{(+0.4)}                         & \multicolumn{1}{l|}{\textbf{62.4}\gain{(+0.4)}} & \textbf{36.9\gain{(+1.1)}}                          & \multicolumn{1}{l|}{\textbf{65.4}\gain{(+2.6)}}       & 57.6\loss{(-0.2)}                             & \multicolumn{1}{l|}{\textbf{31.9}\gain{(+1.9)}} & \textbf{43.7}\gain{(+3.1)}                                & \textbf{86.5}\gain{(+2.0)}                       & \textbf{57.1}                                     \\ \midrule
\end{tabular}}
\label{table:High Resolution}
\end{table*}
\subsection{Method Generalizability}
We validate the generalization and robustness of ASVR under different data scales and LLM backbone configurations, as summarized in Table~\ref{table:Generality}.

\paragraph{The Impact of Data Scaling}
To investigate the effectiveness of ASVR under varying training data scales, we follow Bunny (Bunny-pretrain-LAION-2M \citep{he2024efficientmultimodallearningdatacentric} for pre-training and Bunny-v1\_1-data-2M \citep{he2024efficientmultimodallearningdatacentric} for instruction tuning) and LLaVA-OV~\citep{li2024llavaonevisioneasyvisualtask} (558K for pretraining, 4M for midtraining and 4M for instruction tuning follwing LLaVA-OV ~\citep{li2024llavaonevisioneasyvisualtask} training recipe). As shown in Table~\ref{table: main results} and Table~\ref{table:Generality}, ASVR consistently yields substantial improvements over the baseline across different training data scales, demonstrating its ability to effectively leverage additional data through autoregressive semantic visual reconstruction. 



 


\paragraph{The Impact of LLM Backbone Capacities} We further evaluate the generalizability of ASVR across different LLM backbones. Specifically, we extend our experiments to Mistral-7B\citep{jiang2023mistral} and Vicuna-v1.5-13B, which differ from Vicuna-v1.5-7B in model family and scale~\citep{zheng2023judgingllmasajudgemtbenchchatbot}. As shown in Table~\ref{table:Generality}, ASVR consistently surpasses the baseline across multimodal understanding benchmarks, maintaining strong performance advantages regardless of backbone variations. These results demonstrating both its robustness and adaptability in diverse LLM configurations. The backbone scaling experiment and clear scaling law table will provide in Appendix~\ref{app:scal}.

\subsection{High-resolution Adaptation}
ASVR is also compatible with existing high-resolution strategies and can further enhance the multimodal understanding capabilities of LVLMs. To evaluate the effectiveness of ASVR under high-resolution configurations, we upscale the input resolution of both ASVR and the baseline models to 1152 × 1152, while keeping the training conditions identical. We use LLaVA-558K\citep{liu2023llava} for the pre-training stage and LLaVA-Next-779K\citep{liu2024llava} for instruction tuning following LLaVA-Next settings~\citep{liu2024llava}.
As shown in Table~\ref{table:High Resolution}, under high-resolution configurations, ASVR consistently outperforms the baseline by 2\% in average scores across 14 multimodal benchmarks, further demonstrating its flexibility and robustness across different input resolutions.

\subsection{Comparison with ROSS}
\label{app:ross}
\begin{table*}[th!]
\centering
\caption{\textbf{The detailed comparision between ASVR and ROSS ablation variant. ASVR achieves the best performance under identical training conditions.}
ROSS models visual information through a denoising approach, whereas ASVR adopts unified autoregressive paradigm. The SigLIP-ViT-SO400M/14@384 is utilized for semantic visual supervision and VAE features is appearance visual supervision.}
\tablestyle{5pt}{1.5}
\setlength\tabcolsep{2.2pt}
\resizebox{1\textwidth}{!}{
\begin{tabular}{c|c|c|c|c|cccc|cccc|cc|cc|cc|c}
\toprule
Method & Visual Supervision & {\color[HTML]{333333} \textbf{LLM backbone}} & {\color[HTML]{333333} \textbf{Visual Modeling}} & {\color[HTML]{333333} \textbf{Data}} & {\color[HTML]{333333} \textbf{TVQA}} & {\color[HTML]{333333} \textbf{DVQA}} & {\color[HTML]{333333} \textbf{OCRB}} & {\color[HTML]{333333} \textbf{CQA}}  & {\color[HTML]{333333} \textbf{MMB}}  & {\color[HTML]{333333} \textbf{MME}} & {\color[HTML]{333333} \textbf{SEED}} & {\color[HTML]{333333} \textbf{GQA}}  & {\color[HTML]{333333} \textbf{MMMU}} & {\color[HTML]{333333} \textbf{AI2D}} & {\color[HTML]{333333} \textbf{RQA}}  & {\color[HTML]{333333} \textbf{MMVP}} & {\color[HTML]{333333} \textbf{Hbench}} & {\color[HTML]{333333} \textbf{POPE}} & {\color[HTML]{333333} \textbf{AVG}}  \\ \midrule
ROSS   & Apperance VAE      & {\color[HTML]{333333} Vicuna-v1.5–7B}        & {\color[HTML]{333333} Denoising}                & {\color[HTML]{333333} LLaVA–Next}    & {\color[HTML]{333333} 56.3}          & {\color[HTML]{333333} 39.6}          & {\color[HTML]{333333} 35.9}          & {\color[HTML]{333333} 41.0}            & {\color[HTML]{333333} 65.6}          & {\color[HTML]{333333} 71.7}         & {\color[HTML]{333333} 65.9}          & {\color[HTML]{333333} 61.6}          & {\color[HTML]{333333} 34.4}          & {\color[HTML]{333333} 65.5}          & {\color[HTML]{333333} 55.0}            & {\color[HTML]{333333} 33.3}          & {\color[HTML]{333333} 28.9}            & {\color[HTML]{333333} 85.9}          & {\color[HTML]{333333} 52.9}          \\
ROSS   & Semantic Siglip    & {\color[HTML]{333333} Vicuna-v1.5–7B}        & {\color[HTML]{333333} Denoising}                & {\color[HTML]{333333} LLaVA–Next}    & {\color[HTML]{333333} 57.5}          & {\color[HTML]{333333} 40.2}          & {\color[HTML]{333333} 37.4}          & {\color[HTML]{333333} 42.5}          & {\color[HTML]{333333} 67.0}            & {\color[HTML]{333333} 70.5}         & {\color[HTML]{333333} 66.2}          & {\color[HTML]{333333} 62.1}          & {\color[HTML]{333333} \textbf{34.9}} & {\color[HTML]{333333} 64.6}          & {\color[HTML]{333333} \textbf{55.7}} & {\color[HTML]{333333} 30.1}          & {\color[HTML]{333333} 31.2}            & {\color[HTML]{333333} 85.8}          & {\color[HTML]{333333} 53.3}          \\
\rowcolor[HTML]{E1EAFF}
\textbf{ASVR}   & Semantic Siglip    & {\color[HTML]{333333} Vicuna-v1.5–7B}        & {\color[HTML]{333333} Autoregressive}           & {\color[HTML]{333333} LLaVA–Next}    & {\color[HTML]{333333} \textbf{58.6}} & {\color[HTML]{333333} \textbf{40.6}} & {\color[HTML]{333333} \textbf{39.7}} & {\color[HTML]{333333} \textbf{43.4}} & {\color[HTML]{333333} \textbf{67.9}} & {\color[HTML]{333333} \textbf{73.0}}  & {\color[HTML]{333333} \textbf{67.5}} & {\color[HTML]{333333} \textbf{62.9}} & {\color[HTML]{333333} 34.2}          & {\color[HTML]{333333} \textbf{65.8}} & {\color[HTML]{333333} 55.4}          & {\color[HTML]{333333} \textbf{36.8}} & {\color[HTML]{333333} \textbf{39.2}}   & {\color[HTML]{333333} \textbf{85.9}} & {\color[HTML]{333333} \textbf{55.1}} \\ \hline
\end{tabular}}
\label{ross}
\end{table*}
ROSS~\citep{wang2024ross} reconstructs continuous, appearance-level visual features (VAE features) through denoising, whereas our ASVR reconstructs discrete, semantic-level visual indices (such as discretized SigLIP features) via autoregression. We conduct experiments using the LLaVA-Next dataset~\citep{liu2024llava} under identical training settings, clearly demonstrating that the ASVR-trained model consistently outperforms the ROSS ablation variants across multiple multimodal evaluation metrics. We also implement an additional variant of ROSS that reconstructs continuous semantic-level features (SigLIP features) through denoising. The result is shown in the table below shown in Table~\ref{ross}.

Our ASVR-trained model achieves the best performance, indicating that autoregressive semantic visual reconstruction (ASVR) is superior to both denoising semantic visual reconstruction ablation variants and even denoising appearance visual reconstruction (ROSS) ablation variants. We attribute this performance gap to a fundamental alignment principle: LLMs are inherently trained to model high-level semantic information. Therefore, when the visual supervision is semantically aligned with textual inputs—as in ASVR—it naturally leads to better integration and understanding. In contrast, reconstructing low-level visual features (as in appearance-based ROSS) lacks semantic alignment and can even hinder comprehension. Since tasks such as VQA rely heavily on semantic reasoning, reconstructing semantic visual information is more effective for enhancing multimodal understanding.

\subsection{Qualitative Comparison}
\vspace{-2mm}
\begin{figure}[!h]
    \centering
    \includegraphics[width=0.92\linewidth]{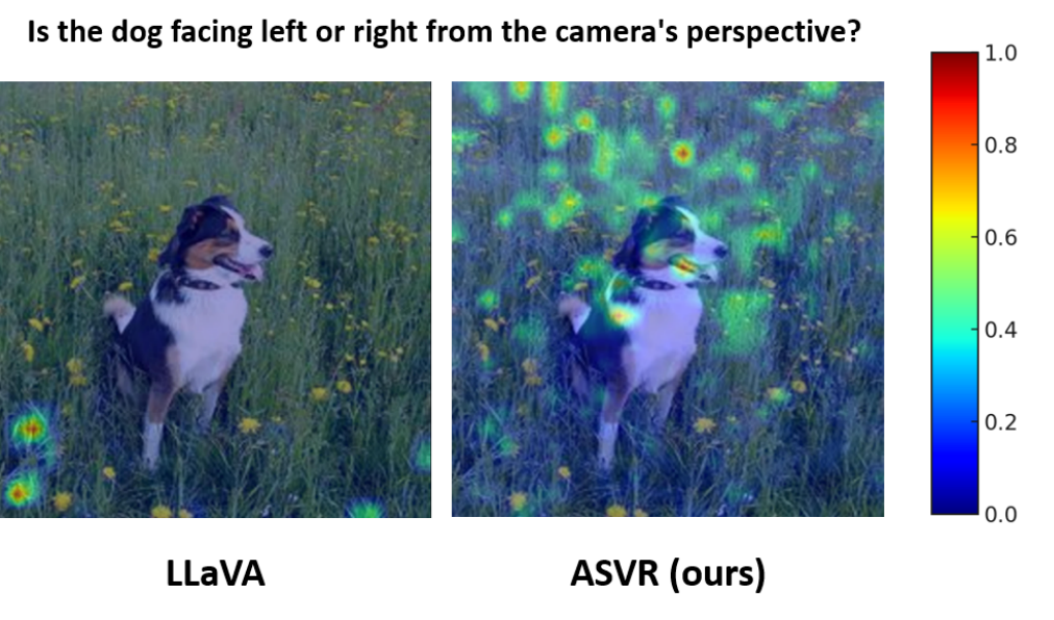} 
    \caption{Qualitative comparison on attention maps, where we keep the same LLM and training data. 
    ASVR urges the model to focus on specific image contents corresponding to the question with higher attention values.}
    \vspace{-2mm}
    \label{fig:attn_1}
\end{figure}
We visualize attention-score maps from several cases, illustrating the attention distribution of the last token with respect to all visual tokens, as shown in Figure~\ref{fig:attn_1}. Compared to the baseline (LLaVA), our ASVR method consistently demonstrates more precise focus on image regions relevant to the given textual query. This highlights that incorporating semantic visual supervision via the autoregressive semantic visual reconstruction objective $\mathcal{L}_{\mathrm{AR}}^{\text{vision}}$ effectively enhance its ability to accurately associate textual descriptions with corresponding visual elements. More comparison on attention maps are shown in Appendix~\ref{more_attn}.



\section{Related Work}
\label{related}
\paragraph{Large Vision Language Models}
The rapid progress in large language models (LLMs)\citep{bai2023qwenlan,llama3modelcard,touvron2023llama2,bi2024deepseek,openai2023gpt4,openai2023chatgpt} has showcased their strong generalization and remarkable instruction-following capabilities. To further expand these strengths for interpreting and interacting with the world
through both visual and linguistic channels. There has been growing interest in Large Vision-Language Models (LVLMs)\citep{liu2023llava,liu2023improved,liu2024llava}, typically trained using a straightforward two-stage visual instruction tuning paradigm~\citep{liu2023llava}, and align visual features extracted by visual encoder with the knowledge and reasoning capabilities of LLMs through the lightweight projector. This process involves jointly training the projector and the LLM on visual instruction datasets, with optional fine-tuning of the visual encoder. However, supervision is limited to text outputs. ASVR introduces a novel autoregressive visual semantic supervision mechanism that encourages the LVLM to reconstruct semantic visual tokens, enhancing its multimodal understanding capabilities.

\paragraph{Visual Autoregression for LVLMs}
Recent approaches~\citep{team2024chameleon,qu2024tokenflow,wang2024emu3,wu2024vilau,wu2024janus} introduce autoregressive visual supervision via visual tokenizers, such as VQGAN~\citep{esser2021vqgan} and VQ-VAE~\citep{oord2018neuraldiscreterepresentationlearning}, enabling LVLMs to support both multimodal understanding and image generation by predict relevant next visual tokens, which are then decoded into images. In contrast, ~ASVR focuses specifically on enhancing the multimodal understanding capability of LVLMs. Rather than generating images, ~ASVR employs autoregressive visual supervision to reconstruct semantic visual tokens within the given continuous image features as input. While prior methods are generative, ~ASVR adopts the reconstructive approach aimed at promoting perception of visual information.

\paragraph{Reconstructive Objectives for LVLMs} ROSS\cite{wang2024reconstructivevisualinstructiontuning} introduces visual supervision for LVLMs by applying denoising objective to reconstruct continuous, appearance-level visual features (VAE features). In contrast, ~ASVR proposes a unified approach by employing autoregressive objective—analogous to that used for text—to reconstruct semantic visual tokens. This design enables seamless integration of visual and textual information under a unified next-token prediction paradigm.

\section{Conclusion}
\label{sec:conclusion}
In summary, we introduced \textbf{Autoregressive Semantic Visual Reconstruction (ASVR)}, enabling joint learning of visual and textual modalities within a unified autoregressive framework and effectively improving multimodal understanding capability of LVLMs. ASVR explicitly integrates semantic visual supervision on visual inputs to foster deep perception. Our findings indicate that autoregressively reconstructing semantic visual representations of images consistently enhances performance across diverse multimodal tasks and also outperform its denoising-based counterpart. This effectiveness is robust across different visual feature types, LLM backbone capacities, data scales, and high-resolution scenarios, underscoring ASVR's adaptability, scalability and versatility. 

\section*{Limitations} 
Due to computational resource constraints, the current implementation of ASVR does not incorporate video understanding capabilities or dynamic resolution functionalities. However, these features are readily transferable and can be integrated in future iterations. In subsequent work, we plan to scale the model to support these functionalities, while also expanding its capabilities to include additional modalities. Furthermore, we aim to extend ASVR’s capacity to enable multimodal generation, thereby broadening its applicability in various domains.
\bibliography{custom}

@inproceedings{esser2021vqgan,
  title={Taming transformers for high-resolution image synthesis},
  author={Esser, Patrick and Rombach, Robin and Ommer, Bjorn},
  booktitle={Proceedings of the IEEE/CVF conference on computer vision and pattern recognition},
  pages={12873--12883},
  year={2021}
}

@article{dosovitskiy2016L2,
  title={Generating images with perceptual similarity metrics based on deep networks},
  author={Dosovitskiy, Alexey and Brox, Thomas},
  journal={Advances in neural information processing systems},
  volume={29},
  year={2016}
}

@inproceedings{zhang2018lpips,
  title={The unreasonable effectiveness of deep features as a perceptual metric},
  author={Zhang, Richard and Isola, Phillip and Efros, Alexei A and Shechtman, Eli and Wang, Oliver},
  booktitle={Proceedings of the IEEE conference on computer vision and pattern recognition},
  pages={586--595},
  year={2018}
}

@inproceedings{isola2017gan,
  title={Image-to-image translation with conditional adversarial networks},
  author={Isola, Phillip and Zhu, Jun-Yan and Zhou, Tinghui and Efros, Alexei A},
  booktitle={Proceedings of the IEEE conference on computer vision and pattern recognition},
  pages={1125--1134},
  year={2017}
}

@inproceedings{radford2021clip,
  title={Learning transferable visual models from natural language supervision},
  author={Radford, Alec and Kim, Jong Wook and Hallacy, Chris and Ramesh, Aditya and Goh, Gabriel and Agarwal, Sandhini and Sastry, Girish and Askell, Amanda and Mishkin, Pamela and Clark, Jack and others},
  booktitle={International conference on machine learning},
  pages={8748--8763},
  year={2021}
}

@inproceedings{zhai2023siglip,
  title={Sigmoid loss for language image pre-training},
  author={Zhai, Xiaohua and Mustafa, Basil and Kolesnikov, Alexander and Beyer, Lucas},
  booktitle={Proceedings of the IEEE/CVF international conference on computer vision},
  pages={11975--11986},
  year={2023}
}

@article{alabdulmohsin2023so400m,
  title={Getting vit in shape: Scaling laws for compute-optimal model design},
  author={Alabdulmohsin, Ibrahim M and Zhai, Xiaohua and Kolesnikov, Alexander and Beyer, Lucas},
  journal={Advances in Neural Information Processing Systems},
  volume={36},
  pages={16406--16425},
  year={2023}
}

@article{liu2023llava,
  title={Visual instruction tuning},
  author={Liu, Haotian and Li, Chunyuan and Wu, Qingyang and Lee, Yong Jae},
  journal={Advances in neural information processing systems},
  volume={36},
  pages={34892--34916},
  year={2023}
}

@inproceedings{liu2024llava1d5,
  title={Improved baselines with visual instruction tuning},
  author={Liu, Haotian and Li, Chunyuan and Li, Yuheng and Lee, Yong Jae},
  booktitle={Proceedings of the IEEE/CVF Conference on Computer Vision and Pattern Recognition},
  pages={26296--26306},
  year={2024}
}

@misc{liu2024llavanext,
    title={LLaVA-NeXT: Improved reasoning, OCR, and world knowledge},
    url={https://llava-vl.github.io/blog/2024-01-30-llava-next/},
    author={Liu, Haotian and Li, Chunyuan and Li, Yuheng and Li, Bo and Zhang, Yuanhan and Shen, Sheng and Lee, Yong Jae},
    month={January},
    year={2024}
}

@misc{Qwen-VL,
  title={Qwen-VL: A Versatile Vision-Language Model for Understanding, Localization, Text Reading, and Beyond},
  author={Bai, Jinze and Bai, Shuai and Yang, Shusheng and Wang, Shijie and Tan, Sinan and Wang, Peng and Lin, Junyang and Zhou, Chang and Zhou, Jingren},
  journal={arXiv preprint arXiv:2308.12966},
  year={2023}
}

@article{Qwen2VL,
  title={Qwen2-VL: Enhancing Vision-Language Model's Perception of the World at Any Resolution},
  author={Wang, Peng and Bai, Shuai and Tan, Sinan and Wang, Shijie and Fan, Zhihao and Bai, Jinze and Chen, Keqin and Liu, Xuejing and Wang, Jialin and Ge, Wenbin and Fan, Yang and Dang, Kai and Du, Mengfei and Ren, Xuancheng and Men, Rui and Liu, Dayiheng and Zhou, Chang and Zhou, Jingren and Lin, Junyang},
  journal={arXiv preprint arXiv:2409.12191},
  year={2024}
}

@inproceedings{chen2024internvl,
  title={Internvl: Scaling up vision foundation models and aligning for generic visual-linguistic tasks},
  author={Chen, Zhe and Wu, Jiannan and Wang, Wenhai and Su, Weijie and Chen, Guo and Xing, Sen and Zhong, Muyan and Zhang, Qinglong and Zhu, Xizhou and Lu, Lewei and others},
  booktitle={Proceedings of the IEEE/CVF Conference on Computer Vision and Pattern Recognition},
  pages={24185--24198},
  year={2024}
}

@misc{lu2024deepseekvl,
      title={DeepSeek-VL: Towards Real-World Vision-Language Understanding},
      author={Haoyu Lu and Wen Liu and Bo Zhang and Bingxuan Wang and Kai Dong and Bo Liu and Jingxiang Sun and Tongzheng Ren and Zhuoshu Li and Hao Yang and Yaofeng Sun and Chengqi Deng and Hanwei Xu and Zhenda Xie and Chong Ruan},
      year={2024},
      eprint={2403.05525},
      archivePrefix={arXiv},
      primaryClass={cs.AI}
}

@misc{wu2024deepseekvl2,
      title={DeepSeek-VL2: Mixture-of-Experts Vision-Language Models for Advanced Multimodal Understanding},
      author={Zhiyu Wu and Xiaokang Chen and Zizheng Pan and Xingchao Liu and Wen Liu and Damai Dai and Huazuo Gao and Yiyang Ma and Chengyue Wu and Bingxuan Wang and Zhenda Xie and Yu Wu and Kai Hu and Jiawei Wang and Yaofeng Sun and Yukun Li and Yishi Piao and Kang Guan and Aixin Liu and Xin Xie and Yuxiang You and Kai Dong and Xingkai Yu and Haowei Zhang and Liang Zhao and Yisong Wang and Chong Ruan},
      year={2024},
      eprint={2412.10302},
      archivePrefix={arXiv},
      primaryClass={cs.CV},
      url={https://arxiv.org/abs/2412.10302},
}

@article{wu2024vilau,
  title={Vila-u: a unified foundation model integrating visual understanding and generation},
  author={Wu, Yecheng and Zhang, Zhuoyang and Chen, Junyu and Tang, Haotian and Li, Dacheng and Fang, Yunhao and Zhu, Ligeng and Xie, Enze and Yin, Hongxu and Yi, Li and others},
  journal={arXiv preprint arXiv:2409.04429},
  year={2024}
}

@article{team2024chameleon,
  title={Chameleon: Mixed-modal early-fusion foundation models},
  author={Team, Chameleon},
  journal={arXiv preprint arXiv:2405.09818},
  year={2024}
}

@article{wang2024emu3,
  title={Emu3: Next-token prediction is all you need},
  author={Wang, Xinlong and Zhang, Xiaosong and Luo, Zhengxiong and Sun, Quan and Cui, Yufeng and Wang, Jinsheng and Zhang, Fan and Wang, Yueze and Li, Zhen and Yu, Qiying and others},
  journal={arXiv preprint arXiv:2409.18869},
  year={2024}
}

@article{wu2024janus,
  title={Janus: Decoupling visual encoding for unified multimodal understanding and generation},
  author={Wu, Chengyue and Chen, Xiaokang and Wu, Zhiyu and Ma, Yiyang and Liu, Xingchao and Pan, Zizheng and Liu, Wen and Xie, Zhenda and Yu, Xingkai and Ruan, Chong and others},
  journal={arXiv preprint arXiv:2410.13848},
  year={2024}
}

@article{xie2024musevl,
  title={MUSE-VL: Modeling Unified VLM through Semantic Discrete Encoding},
  author={Xie, Rongchang and Du, Chen and Song, Ping and Liu, Chang},
  journal={arXiv preprint arXiv:2411.17762},
  year={2024}
}

@article{qu2024tokenflow,
  title={Tokenflow: Unified image tokenizer for multimodal understanding and generation},
  author={Qu, Liao and Zhang, Huichao and Liu, Yiheng and Wang, Xu and Jiang, Yi and Gao, Yiming and Ye, Hu and Du, Daniel K and Yuan, Zehuan and Wu, Xinglong},
  journal={arXiv preprint arXiv:2412.03069},
  year={2024}
}

@article{metamorph,
  title={MetaMorph: Multimodal Understanding and Generation via Instruction Tuning},
  author={Tong, Shengbang and Fan, David and Zhu, Jiachen and Xiong, Yunyang and Chen, Xinlei and Sinha, Koustuv and Rabbat, Michael and LeCun, Yann and Xie, Saining and Liu, Zhuang},
  journal={arXiv preprint arXiv:2412.14164},
  year={2024}
}

@article{liu2023mmbench,
  title={Mmbench: Is your multi-modal model an all-around player?},
  author={Liu, Yuan and Duan, Haodong and Zhang, Yuanhan and Li, Bo and Zhang, Songyang and Zhao, Wangbo and Yuan, Yike and Wang, Jiaqi and He, Conghui and Liu, Ziwei and others},
  journal={arXiv preprint arXiv:2307.06281},
  year={2023}
}

@misc{fu2024mme,
      title={MME: A Comprehensive Evaluation Benchmark for Multimodal Large Language Models}, 
      author={Chaoyou Fu and Peixian Chen and Yunhang Shen and Yulei Qin and Mengdan Zhang and Xu Lin and Jinrui Yang and Xiawu Zheng and Ke Li and Xing Sun and Yunsheng Wu and Rongrong Ji},
      year={2024},
      eprint={2306.13394},
      archivePrefix={arXiv},
      primaryClass={cs.CV}
}

@article{yue2023mmmu,
  title={Mmmu: A massive multi-discipline multimodal understanding and reasoning benchmark for expert agi},
  author={Yue, Xiang and Ni, Yuansheng and Zhang, Kai and Zheng, Tianyu and Liu, Ruoqi and Zhang, Ge and Stevens, Samuel and Jiang, Dongfu and Ren, Weiming and Sun, Yuxuan and others},
  journal={arXiv preprint arXiv:2311.16502},
  year={2023}
}

@InProceedings{vqa_v2,
author = {Yash Goyal and Tejas Khot and Douglas Summers{-}Stay and Dhruv Batra and Devi Parikh},
title = {Making the {V} in {VQA} Matter: Elevating the Role of Image Understanding in {V}isual {Q}uestion {A}nswering},
booktitle = {Conference on Computer Vision and Pattern Recognition (CVPR)},
year = {2017},
}

@inproceedings{gqa,
  title={Gqa: A new dataset for real-world visual reasoning and compositional question answering},
  author={Hudson, Drew A and Manning, Christopher D},
  booktitle={Proceedings of the IEEE/CVF conference on computer vision and pattern recognition},
  pages={6700--6709},
  year={2019}
}

@article{POPE,
  title={Evaluating object hallucination in large vision-language models},
  author={Li, Yifan and Du, Yifan and Zhou, Kun and Wang, Jinpeng and Zhao, Wayne Xin and Wen, Ji-Rong},
  journal={arXiv preprint arXiv:2305.10355},
  year={2023}
}

@inproceedings{deng2009imagenet,
  title={Imagenet: A large-scale hierarchical image database},
  author={Deng, Jia and Dong, Wei and Socher, Richard and Li, Li-Jia and Li, Kai and Fei-Fei, Li},
  booktitle={2009 IEEE conference on computer vision and pattern recognition},
  pages={248--255},
  year={2009},
  organization={Ieee}
}

@article{wang2024ross,
  title={Reconstructive visual instruction tuning},
  author={Wang, Haochen and Zheng, Anlin and Zhao, Yucheng and Wang, Tiancai and Ge, Zheng and Zhang, Xiangyu and Zhang, Zhaoxiang},
  journal={arXiv preprint arXiv:2410.09575},
  year={2024}
}

@misc{li2024llavaonevisioneasyvisualtask,
      title={LLaVA-OneVision: Easy Visual Task Transfer}, 
      author={Bo Li and Yuanhan Zhang and Dong Guo and Renrui Zhang and Feng Li and Hao Zhang and Kaichen Zhang and Peiyuan Zhang and Yanwei Li and Ziwei Liu and Chunyuan Li},
      year={2024},
      eprint={2408.03326},
      archivePrefix={arXiv},
      primaryClass={cs.CV},
      url={https://arxiv.org/abs/2408.03326}, 
}

@misc{dong2024internlmxcomposer24khdpioneeringlargevisionlanguage,
      title={InternLM-XComposer2-4KHD: A Pioneering Large Vision-Language Model Handling Resolutions from 336 Pixels to 4K HD}, 
      author={Xiaoyi Dong and Pan Zhang and Yuhang Zang and Yuhang Cao and Bin Wang and Linke Ouyang and Songyang Zhang and Haodong Duan and Wenwei Zhang and Yining Li and Hang Yan and Yang Gao and Zhe Chen and Xinyue Zhang and Wei Li and Jingwen Li and Wenhai Wang and Kai Chen and Conghui He and Xingcheng Zhang and Jifeng Dai and Yu Qiao and Dahua Lin and Jiaqi Wang},
      year={2024},
      eprint={2404.06512},
      archivePrefix={arXiv},
      primaryClass={cs.CV},
      url={https://arxiv.org/abs/2404.06512}, 
}

@misc{wang2024qwen2vlenhancingvisionlanguagemodels,
      title={Qwen2-VL: Enhancing Vision-Language Model's Perception of the World at Any Resolution}, 
      author={Peng Wang and Shuai Bai and Sinan Tan and Shijie Wang and Zhihao Fan and Jinze Bai and Keqin Chen and Xuejing Liu and Jialin Wang and Wenbin Ge and Yang Fan and Kai Dang and Mengfei Du and Xuancheng Ren and Rui Men and Dayiheng Liu and Chang Zhou and Jingren Zhou and Junyang Lin},
      year={2024},
      eprint={2409.12191},
      archivePrefix={arXiv},
      primaryClass={cs.CV},
      url={https://arxiv.org/abs/2409.12191}, 
}

@misc{liu2024llava,
  title={Llava-next: Improved reasoning, ocr, and world knowledge},
  author={Liu, Haotian and Li, Chunyuan and Li, Yuheng and Li, Bo and Zhang, Yuanhan and Shen, Sheng and Lee, Yong Jae},
  year={2024}
}

@article{yao2024minicpm,
  title={MiniCPM-V: A GPT-4V Level MLLM on Your Phone},
  author={Yao, Yuan and Yu, Tianyu and Zhang, Ao and Wang, Chongyi and Cui, Junbo and Zhu, Hongji and Cai, Tianchi and Li, Haoyu and Zhao, Weilin and He, Zhihui and others},
  journal={arXiv preprint arXiv:2408.01800},
  year={2024}
}

@misc{tong2024cambrian1fullyopenvisioncentric,
      title={Cambrian-1: A Fully Open, Vision-Centric Exploration of Multimodal LLMs}, 
      author={Shengbang Tong and Ellis Brown and Penghao Wu and Sanghyun Woo and Manoj Middepogu and Sai Charitha Akula and Jihan Yang and Shusheng Yang and Adithya Iyer and Xichen Pan and Ziteng Wang and Rob Fergus and Yann LeCun and Saining Xie},
      year={2024},
      eprint={2406.16860},
      archivePrefix={arXiv},
      primaryClass={cs.CV},
      url={https://arxiv.org/abs/2406.16860}, 
}

@article{song2025dualtoken,
  title={Dualtoken: Towards unifying visual understanding and generation with dual visual vocabularies},
  author={Song, Wei and Wang, Yuran and Song, Zijia and Li, Yadong and Sun, Haoze and Chen, Weipeng and Zhou, Zenan and Xu, Jianhua and Wang, Jiaqi and Yu, Kaicheng},
  journal={arXiv preprint arXiv:2503.14324},
  year={2025}
}

@misc{zheng2023judgingllmasajudgemtbenchchatbot,
      title={Judging LLM-as-a-Judge with MT-Bench and Chatbot Arena}, 
      author={Lianmin Zheng and Wei-Lin Chiang and Ying Sheng and Siyuan Zhuang and Zhanghao Wu and Yonghao Zhuang and Zi Lin and Zhuohan Li and Dacheng Li and Eric P. Xing and Hao Zhang and Joseph E. Gonzalez and Ion Stoica},
      year={2023},
      eprint={2306.05685},
      archivePrefix={arXiv},
      primaryClass={cs.CL},
      url={https://arxiv.org/abs/2306.05685}, 
}

@misc{liu2024mmbenchmultimodalmodelallaround,
      title={MMBench: Is Your Multi-modal Model an All-around Player?}, 
      author={Yuan Liu and Haodong Duan and Yuanhan Zhang and Bo Li and Songyang Zhang and Wangbo Zhao and Yike Yuan and Jiaqi Wang and Conghui He and Ziwei Liu and Kai Chen and Dahua Lin},
      year={2024},
      eprint={2307.06281},
      archivePrefix={arXiv},
      primaryClass={cs.CV},
      url={https://arxiv.org/abs/2307.06281}, 
}

@misc{hudson2019gqanewdatasetrealworld,
      title={GQA: A New Dataset for Real-World Visual Reasoning and Compositional Question Answering}, 
      author={Drew A. Hudson and Christopher D. Manning},
      year={2019},
      eprint={1902.09506},
      archivePrefix={arXiv},
      primaryClass={cs.CL},
      url={https://arxiv.org/abs/1902.09506}, 
}

@misc{li2023seedbenchbenchmarkingmultimodalllms,
      title={SEED-Bench: Benchmarking Multimodal LLMs with Generative Comprehension}, 
      author={Bohao Li and Rui Wang and Guangzhi Wang and Yuying Ge and Yixiao Ge and Ying Shan},
      year={2023},
      eprint={2307.16125},
      archivePrefix={arXiv},
      primaryClass={cs.CL},
      url={https://arxiv.org/abs/2307.16125}, 
}

@misc{fu2024mmecomprehensiveevaluationbenchmark,
      title={MME: A Comprehensive Evaluation Benchmark for Multimodal Large Language Models}, 
      author={Chaoyou Fu and Peixian Chen and Yunhang Shen and Yulei Qin and Mengdan Zhang and Xu Lin and Jinrui Yang and Xiawu Zheng and Ke Li and Xing Sun and Yunsheng Wu and Rongrong Ji},
      year={2024},
      eprint={2306.13394},
      archivePrefix={arXiv},
      primaryClass={cs.CV},
      url={https://arxiv.org/abs/2306.13394}, 
}

@misc{singh2019vqamodelsread,
      title={Towards VQA Models That Can Read}, 
      author={Amanpreet Singh and Vivek Natarajan and Meet Shah and Yu Jiang and Xinlei Chen and Dhruv Batra and Devi Parikh and Marcus Rohrbach},
      year={2019},
      eprint={1904.08920},
      archivePrefix={arXiv},
      primaryClass={cs.CL},
      url={https://arxiv.org/abs/1904.08920}, 
}

@misc{masry2022chartqabenchmarkquestionanswering,
      title={ChartQA: A Benchmark for Question Answering about Charts with Visual and Logical Reasoning}, 
      author={Ahmed Masry and Do Xuan Long and Jia Qing Tan and Shafiq Joty and Enamul Hoque},
      year={2022},
      eprint={2203.10244},
      archivePrefix={arXiv},
      primaryClass={cs.CL},
      url={https://arxiv.org/abs/2203.10244}, 
}

@misc{mathew2021docvqadatasetvqadocument,
      title={DocVQA: A Dataset for VQA on Document Images}, 
      author={Minesh Mathew and Dimosthenis Karatzas and C. V. Jawahar},
      year={2021},
      eprint={2007.00398},
      archivePrefix={arXiv},
      primaryClass={cs.CV},
      url={https://arxiv.org/abs/2007.00398}, 
}

@article{Liu_2024,
    title={OCRBench: on the hidden mystery of OCR in large multimodal models},
    volume={67},
    ISSN={1869-1919},
    url={http://dx.doi.org/10.1007/s11432-024-4235-6},
    DOI={10.1007/s11432-024-4235-6},
    number={12},
    journal={Science China Information Sciences},
    publisher={Springer Science and Business Media LLC},
    author={Liu, Yuliang and Li, Zhang and Huang, Mingxin and Yang, Biao and Yu, Wenwen and Li, Chunyuan and Yin, Xu-Cheng and Liu, Cheng-Lin and Jin, Lianwen and Bai, Xiang},
    year={2024},
    month=dec }

@misc{yue2024mmmumassivemultidisciplinemultimodal,
      title={MMMU: A Massive Multi-discipline Multimodal Understanding and Reasoning Benchmark for Expert AGI}, 
      author={Xiang Yue and Yuansheng Ni and Kai Zhang and Tianyu Zheng and Ruoqi Liu and Ge Zhang and Samuel Stevens and Dongfu Jiang and Weiming Ren and Yuxuan Sun and Cong Wei and Botao Yu and Ruibin Yuan and Renliang Sun and Ming Yin and Boyuan Zheng and Zhenzhu Yang and Yibo Liu and Wenhao Huang and Huan Sun and Yu Su and Wenhu Chen},
      year={2024},
      eprint={2311.16502},
      archivePrefix={arXiv},
      primaryClass={cs.CL},
      url={https://arxiv.org/abs/2311.16502}, 
}

@misc{kembhavi2016diagramworthdozenimages,
      title={A Diagram Is Worth A Dozen Images}, 
      author={Aniruddha Kembhavi and Mike Salvato and Eric Kolve and Minjoon Seo and Hannaneh Hajishirzi and Ali Farhadi},
      year={2016},
      eprint={1603.07396},
      archivePrefix={arXiv},
      primaryClass={cs.CV},
      url={https://arxiv.org/abs/1603.07396}, 
}

@misc{li2023evaluatingobjecthallucinationlarge,
      title={Evaluating Object Hallucination in Large Vision-Language Models}, 
      author={Yifan Li and Yifan Du and Kun Zhou and Jinpeng Wang and Wayne Xin Zhao and Ji-Rong Wen},
      year={2023},
      eprint={2305.10355},
      archivePrefix={arXiv},
      primaryClass={cs.CV},
      url={https://arxiv.org/abs/2305.10355}, 
}

@misc{guan2024hallusionbenchadvanceddiagnosticsuite,
      title={HallusionBench: An Advanced Diagnostic Suite for Entangled Language Hallucination and Visual Illusion in Large Vision-Language Models}, 
      author={Tianrui Guan and Fuxiao Liu and Xiyang Wu and Ruiqi Xian and Zongxia Li and Xiaoyu Liu and Xijun Wang and Lichang Chen and Furong Huang and Yaser Yacoob and Dinesh Manocha and Tianyi Zhou},
      year={2024},
      eprint={2310.14566},
      archivePrefix={arXiv},
      primaryClass={cs.CV},
      url={https://arxiv.org/abs/2310.14566}, 
}

@misc{tong2024eyeswideshutexploring,
      title={Eyes Wide Shut? Exploring the Visual Shortcomings of Multimodal LLMs}, 
      author={Shengbang Tong and Zhuang Liu and Yuexiang Zhai and Yi Ma and Yann LeCun and Saining Xie},
      year={2024},
      eprint={2401.06209},
      archivePrefix={arXiv},
      primaryClass={cs.CV},
      url={https://arxiv.org/abs/2401.06209}, 
}

@misc{xai2024grok,
  author       = {{xAI}},
  title        = {Grok},
  year         = {2024},
  note         = {Developed by xAI},
  howpublished = {\url{https://x.ai}}
}

@misc{he2024efficientmultimodallearningdatacentric,
      title={Efficient Multimodal Learning from Data-centric Perspective}, 
      author={Muyang He and Yexin Liu and Boya Wu and Jianhao Yuan and Yueze Wang and Tiejun Huang and Bo Zhao},
      year={2024},
      eprint={2402.11530},
      archivePrefix={arXiv},
      primaryClass={cs.CV},
      url={https://arxiv.org/abs/2402.11530}, 
}

@article{llama3modelcard,

title={Llama 3 Model Card},

author={AI@Meta},

year={2024},

url = {https://github.com/meta-llama/llama3/blob/main/MODEL_CARD.md}

}

@article{jiang2023mistral,
  title={Mistral 7B},
  author={Jiang, Albert Q and Sablayrolles, Alexandre and Mensch, Arthur and Bamford, Chris and Chaplot, Devendra Singh and Casas, Diego de las and Bressand, Florian and Lengyel, Gianna and Lample, Guillaume and Saulnier, Lucile and others},
  journal={arXiv preprint arXiv:2310.06825},
  year={2023}
}

@misc{dosovitskiy2021imageworth16x16words,
      title={An Image is Worth 16x16 Words: Transformers for Image Recognition at Scale}, 
      author={Alexey Dosovitskiy and Lucas Beyer and Alexander Kolesnikov and Dirk Weissenborn and Xiaohua Zhai and Thomas Unterthiner and Mostafa Dehghani and Matthias Minderer and Georg Heigold and Sylvain Gelly and Jakob Uszkoreit and Neil Houlsby},
      year={2021},
      eprint={2010.11929},
      archivePrefix={arXiv},
      primaryClass={cs.CV},
      url={https://arxiv.org/abs/2010.11929}, 
}

@misc{wang2024reconstructivevisualinstructiontuning,
      title={Reconstructive Visual Instruction Tuning}, 
      author={Haochen Wang and Anlin Zheng and Yucheng Zhao and Tiancai Wang and Zheng Ge and Xiangyu Zhang and Zhaoxiang Zhang},
      year={2024},
      eprint={2410.09575},
      archivePrefix={arXiv},
      primaryClass={cs.CV},
      url={https://arxiv.org/abs/2410.09575}, 
}

@article{bai2023qwenlan,
  title={Qwen technical report},
  author={Bai, Jinze and Bai, Shuai and Chu, Yunfei and Cui, Zeyu and Dang, Kai and Deng, Xiaodong and Fan, Yang and Ge, Wenbin and Han, Yu and Huang, Fei and others},
  journal={arXiv preprint arXiv:2309.16609},
  year={2023}
}

@article{touvron2023llama2,
  title={Llama 2: Open foundation and fine-tuned chat models},
  author={Touvron, Hugo and Martin, Louis and Stone, Kevin and Albert, Peter and Almahairi, Amjad and Babaei, Yasmine and Bashlykov, Nikolay and Batra, Soumya and Bhargava, Prajjwal and Bhosale, Shruti and others},
  journal={arXiv:2307.09288
        
        },
  year={2023}
}

@article{bi2024deepseek,
  title={Deepseek llm: Scaling open-source language models with longtermism},
  author={Bi, Xiao and Chen, Deli and Chen, Guanting and Chen, Shanhuang and Dai, Damai and Deng, Chengqi and Ding, Honghui and Dong, Kai and Du, Qiushi and Fu, Zhe and others},
  journal={arXiv preprint arXiv:2401.02954
        
        },
  year={2024}
}

@article{openai2023gpt4,
  title={GPT-4 Technical Report},
  author={OpenAI},
  journal={arXiv:2303.08774
        
        },
  year={2023}
}

@misc{openai2023chatgpt,
title={ChatGPT (August 3 version)},
author={OpenAI},
year={2023},
url={https://chat.openai.com/chat}
}

@article{liu2023improved,
  title={Improved baselines with visual instruction tuning},
  author={Liu, Haotian and Li, Chunyuan and Li, Yuheng and Lee, Yong Jae},
  journal={arXiv:2310.03744
        
        },
  year={2023}
}

@misc{oord2018neuraldiscreterepresentationlearning,
      title={Neural Discrete Representation Learning}, 
      author={Aaron van den Oord and Oriol Vinyals and Koray Kavukcuoglu},
      year={2018},
      eprint={1711.00937},
      archivePrefix={arXiv},
      primaryClass={cs.LG},
      url={https://arxiv.org/abs/1711.00937}, 
}

@misc{tschannen2025siglip2multilingualvisionlanguage,
      title={SigLIP 2: Multilingual Vision-Language Encoders with Improved Semantic Understanding, Localization, and Dense Features}, 
      author={Michael Tschannen and Alexey Gritsenko and Xiao Wang and Muhammad Ferjad Naeem and Ibrahim Alabdulmohsin and Nikhil Parthasarathy and Talfan Evans and Lucas Beyer and Ye Xia and Basil Mustafa and Olivier Hénaff and Jeremiah Harmsen and Andreas Steiner and Xiaohua Zhai},
      year={2025},
      eprint={2502.14786},
      archivePrefix={arXiv},
      primaryClass={cs.CV},
      url={https://arxiv.org/abs/2502.14786}, 
}

\appendix

\section{Appendix}
\subsection{Use of LLMs in in Paper Writing} In preparing this paper, Large Language Models (LLMs) were employed to support the refinement of writing. Their role was limited to enhancing the linguistic presentation of the paper by improving readability, clarity, and stylistic consistency. Specifically, the models were used for tasks such as rephrasing sentences, checking grammar, and streamlining the flow of the text. We emphasize that the LLMs were not involved in generating research ideas, designing methodologies, or conducting experiments. All conceptual development, methodological design, and analytical work were carried out solely by the authors. The contribution of the LLMs was restricted to language-level improvements and did not extend to the scientific substance of the work. The authors retain complete responsibility for the content of this paper, including passages revised with LLM assistance. Care has been taken to ensure that the use of LLMs complies with ethical standards and does not give rise to plagiarism or any form of scientific misconduct.

\subsection{Qualitative Comparison}
\label{more_attn}
\begin{figure*}[h]
    \centering
    \includegraphics[width=\linewidth]{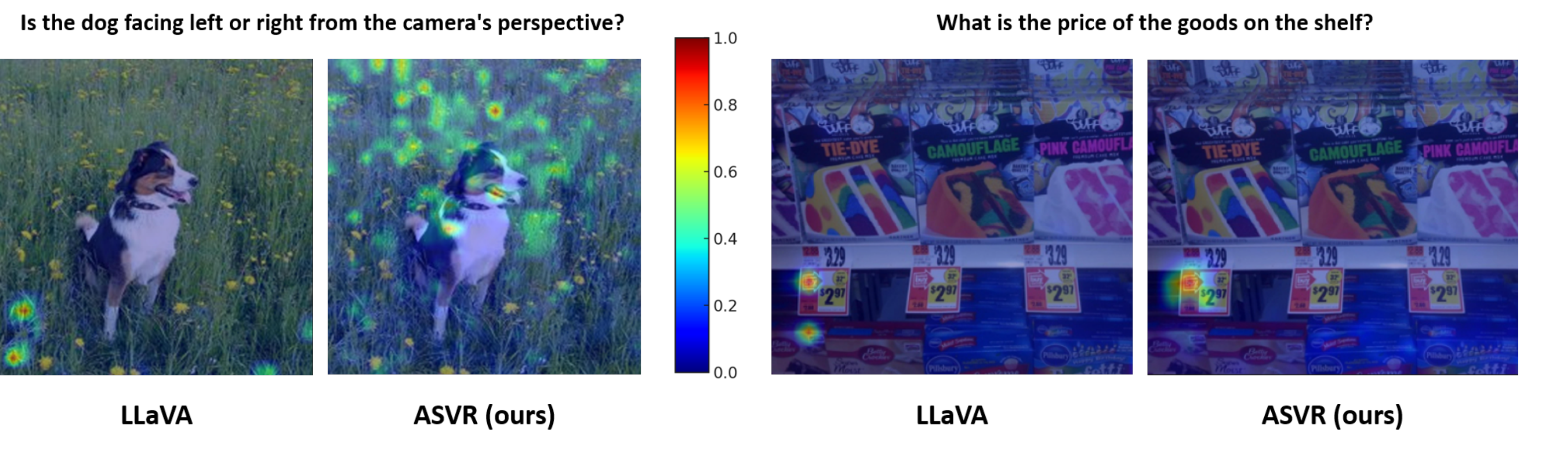} 
    \caption{More qualitative comparison on attention maps, where we keep the same LLM and training data. With extra vision-centric supervision signals, ASVR urges the model to focus on specific image contents corresponding to the question with higher attention values.}
    \label{fig:attn}
\end{figure*}

We visualize attention-score maps from several cases, illustrating the attention distribution of the last token with respect to all visual tokens, as shown in Figure~\ref{fig:attn}. Compared to the baseline (LLaVA), our ASVR method consistently demonstrates more precise focus on image regions relevant to the given textual query. This highlights that incorporating semantic visual supervision via the autoregressive semantic visual reconstruction objective $\mathcal{L}_{\mathrm{AR}}^{\text{vision}}$ effectively enhance its ability to accurately associate textual descriptions with corresponding visual elements.

\subsection{Evaluation Prompts}
\label{eval_prompts}
All prompts used for evaluation benchmarks are released and summarized in Table\ref{table:eval_prompt} following Cambrian-1~\citep{tong2024cambrian1fullyopenvisioncentric}.

\begin{table*}[h]
\centering
\caption{\textbf{Listing the prompts used in the evaluation of each benchmark.}}
\resizebox{0.8\textwidth}{!}{
\begin{tabular}{c|c}
\toprule
Benchmark      & Prompt                                                                              \\ \midrule
TextVQA~\citep{singh2019vqamodelsread}        & Answer the question using a single word or phrase.                                  \\
DocVQA~\citep{mathew2021docvqadatasetvqadocument}         & Answer the question using a single word or phrase.                                  \\
OCRBench~\citep{Liu_2024}       & Give the short answer directly.                                                     \\
ChartQA~\citep{masry2022chartqabenchmarkquestionanswering}        & Answer the question using a single number or phrase.                                \\
MMBench~\citep{liu2024mmbenchmultimodalmodelallaround}        & Answer with the option’s letter from the given choices directly.                    \\
MME~\citep{fu2024mmecomprehensiveevaluationbenchmark}            & Answer the question using a single word or phrase.                                  \\
SEED-Image~\citep{li2023seedbenchbenchmarkingmultimodalllms}     & Answer with the option's letter from the given choices directly.                    \\
GQA~\citep{hudson2019gqanewdatasetrealworld}            & Answer the question using a single word or phrase.                                  \\
MMMU~\citep{yue2024mmmumassivemultidisciplinemultimodal}           & Answer with the option’s letter from the given choices directly.                    \\
AI2D~\citep{kembhavi2016diagramworthdozenimages}           & Answer with the option's letter from the given choices directly.                    \\
RealworldQA~\citep{xai2024grok}    & Please answer directly with only the letter of the correct option and nothing else. \\
MMVP~\citep{tong2024eyeswideshutexploring}           & Answer with the option's letter from the given choices directly.                    \\
Hallusionbench~\citep{guan2024hallusionbenchadvanceddiagnosticsuite} & Answer the question using a single word or phrase.                                  \\
POPE~\citep{li2023evaluatingobjecthallucinationlarge}           & Answer the question using a single word or phrase.                                  \\ \midrule
\end{tabular}}
\label{table:eval_prompt}
\end{table*}

\subsection{The Scalability of ASVR}
\label{app:scal}
we show the clear scaling study along two axes in Table~\ref{data_scaling_tabel} and Table~\ref{model_scaling_tabel}:

\textbf{Data scaling} We train on four datasets—LLaVA-1.5-665K~\citep{liu2023llava}, LLaVA-Next-779K~\citep{liu2024llavanext}, Bunny-2M~\citep{he2024efficientmultimodallearningdatacentric}, and LLaVA-OV-4M~\citep{li2024llavaonevisioneasyvisualtask} to isolate the effect of data volume.
\begin{table}[]
\centering
\caption{\textbf{The scaling relationship between computational cost (FLOPs) and average performance score across different scale dataasets with the same LLM backbone-vicuna-v1.5-7B.}}
\resizebox{0.5\textwidth}{!}{
\begin{tabular}{c|cc}
\toprule
\textbf{Data}           &\textbf{FLOPs (×1e19)} & \textbf{Avg Score} \\ \midrule
LLaVA-1.5-665K~\citep{liu2023llava}  & 1.53          & 49.8      \\ \midrule
LLaVA-Next-779K~\citep{liu2024llavanext} & 2.49          & 55.1      \\ \midrule
Bunny-2M~\citep{he2024efficientmultimodallearningdatacentric}        & 4.52          & 55.7      \\ \midrule
LLaVA-OV-4M~\citep{li2024llavaonevisioneasyvisualtask}   & 7.51          & 57.9      \\ \midrule
\end{tabular}}
\label{data_scaling_tabel}
\end{table}

\textbf{Backbone scaling} Using the same Vicuna family, we vary only the parameter count and scale up to a 13B model (the maximum allowed by our computational budget).

\begin{table}[h!]
\centering
\caption{\textbf{The scaling relationship between computational cost (FLOPs) and average performance score across different scale backbone parameters with the same training dataset-LLaVA-1.5-665k.}}
\resizebox{0.5\textwidth}{!}{
\begin{tabular}{c|cc}
\toprule
{\color[HTML]{333333} \textbf{LLM Backbone}} & {\color[HTML]{333333} \textbf{FLOPs (×1e19)}} & {\color[HTML]{333333} \textbf{Avg Score}} \\ \hline
{\color[HTML]{333333} vicuna-v1.5-7B}         & {\color[HTML]{333333} 1.53}                   & {\color[HTML]{333333} 49.8}               \\ \midrule
{\color[HTML]{333333} vicuna-v1.5-13B}        & {\color[HTML]{333333} 2.58}                   & {\color[HTML]{333333} 52.4}               \\ \midrule
\end{tabular}}
\label{model_scaling_tabel}
\end{table}

\subsection{Ablation Study}
\label{ablation}
\begin{table*}[th!]
\centering
\caption{\textbf{Ablation study for various ASVR configurations.} This table presents a comparison of various ASVR settings, including semantic tokenizer, varied the degree of alignment with text, and the training strategy, where "PT/IT" denotes that semantic visual supervision is applied during both the pre-training and instruction tuning stages, while "IT" indicates that semantic visual supervision is applied only during instruction tuning.}
\tablestyle{5pt}{1.5}
\setlength\tabcolsep{2.2pt}
\resizebox{1\textwidth}{!}{
\begin{tabular}{ccc|llllllllllllll|c}
\toprule
                                  &                            &                                   & \multicolumn{4}{c}{OCR}                                                                                             & \multicolumn{4}{c}{General}                                                                                       & \multicolumn{2}{c}{Knowledge}                               & \multicolumn{2}{c}{Visual-centric}                           & \multicolumn{2}{c|}{Hallusion}                         &                       \\ \cline{4-17}
\multirow{-2}{*}{Ablated Aspects} & \multirow{-2}{*}{Original} & \multirow{-2}{*}{Ablated Setting} & \multicolumn{1}{c}{TVQA} & \multicolumn{1}{c}{DVQA} & \multicolumn{1}{c}{OCRB} & \multicolumn{1}{c|}{CQA}           & \multicolumn{1}{c}{MMB} & \multicolumn{1}{c}{MME} & \multicolumn{1}{c}{SEED} & \multicolumn{1}{c|}{GQA}           & \multicolumn{1}{c}{MMMU} & \multicolumn{1}{c|}{AI2D}        & \multicolumn{1}{c}{RQA} & \multicolumn{1}{c|}{MMVP}          & \multicolumn{1}{c}{HBench} & \multicolumn{1}{c|}{POPE} & \multirow{-2}{*}{AVG} \\ \midrule
Semantic Tokenizer                & 12M              & 3M                      & 57.8\loss{(-1.7)}                     & \textbf{25.4}\gain{(+1.1)}             & 33.1\loss{(-2.3)}                     & \multicolumn{1}{l|}{16.2\loss{(-0.2)}}           & \textbf{67.2}\gain{(+1.1)}           & 70.3\loss{(-2.5)}                    & 64.8\loss{(-1.6)}                      & \multicolumn{1}{l|}{60.0\loss{(-1.5)}}            & 31.8\loss{(-2.1)}                     & \multicolumn{1}{l|}{55.9\loss{(-1.1)}}        & \textbf{54.3\gain{(+0.2)}}           & \multicolumn{1}{l|}{24.7\loss{(-5.3)}}          & 33.0\loss{(-0.7)}                         & 86.1\loss{(-0.2)}                      & 48.6                  \\ \midrule
Training Strategy                & PT/IT                      & IT                                & 55.3\loss{(-4.2)}                     & 18.9\loss{(-5.4)}                     & 29.5\loss{(-5.9)}                     & \multicolumn{1}{l|}{14.0\loss{(-2.4)}}            & 61.2\loss{(-4.9)}                    & 67.8\loss{(-5.0)}                    & 60.5\loss{(-5.9)}                     & \multicolumn{1}{l|}{58.3\loss{(-3.2)}}          & 33.4\loss{(-0.5)}                     & \multicolumn{1}{l|}{52.6\loss{(-4.4)}}        & 52.3\loss{(-1.8)}                    & \multicolumn{1}{l|}{20.8\loss{(-9.2)}}          & 30.0\loss{(-3.7)}                         & 84.9\loss{(-1.4)}                      & 45.7                  \\ \midrule
\rowcolor[HTML]{E1EAFF}

ASVR      & \cellcolor[HTML]{E1EAFF}-  & \cellcolor[HTML]{E1EAFF}-         & \textbf{59.5}            & 24.3                     & \textbf{35.4}            & \multicolumn{1}{l|}{\textbf{16.4}} & 66.1                    & \textbf{72.8}           & \textbf{66.4}            & \multicolumn{1}{l|}{\textbf{61.5}} & \textbf{33.9}            & \multicolumn{1}{l|}{\textbf{57.0}} & 54.1                    & \multicolumn{1}{l|}{\textbf{30.0}} & \textbf{33.7}              & \textbf{86.3}             & \textbf{49.8}         \\ \midrule
\end{tabular}}
\label{table:Ablation Study}
\end{table*}

\paragraph{The Impact of Semantic Tokenizer}
Increasing the degree of alignment with text for semantic tokenizer leads to performance of ASVR. We use VQ-SigLip~\citep{song2025dualtoken} trained on different data scales to construct semantic visual supervision targets: with 3M data trained varient, which achieves zero-shot ImageNet classification accuracy of 78.6\%~\citep{deng2009imagenet}, and with 12M data trained varient, which achieves 81.6\% and thus exhibits stronger semantic alignment. As shown in Table~\ref{table:Ablation Study}, ASVR equipped with the better-aligned with 12M data trained varient consistently outperforms the variant using 3M data trained across the majority of multimodal benchmarks, with the average performance improving by more than 2\%. These results demonstrate that employing better semantically aligned visual tokenizer provides semantic visual supervision targets with more meaningful aspects of the image, and further support our claim that Semantic Visual Reconstruction plays a key role in enhancing the multimodal understanding capabilities of LVLMs. Moreover, when the supervised visual tokenizer provides richer semantic information, ASVR achieves stronger performance. We present the results obtained using discrete SigLIP2~\citep{tschannen2025siglip2multilingualvisionlanguage} as visual tokenizer which contain richer semantic visual information in the Appendix~\ref{app:supervision}.

\paragraph{The Impact of Training Strategy}
We explore different training strategies for ASVR, comparing whether to apply semantic visual supervision in both the pre-training and instruction tuning stages, or to apply it only during instruction tuning, while keeping the pre-training stage purely with text-based autoregressive training. As shown in Table~\ref{table:Ablation Study}, incorporating semantic visual supervision to support visual autoregressive training in both the pre-training and instruction tuning stages consistently outperforms the single-stage variant across all benchmarks, achieving an average performance gain of nearly 6\%. This further underscores the importance of Semantic Visual Reconstruction during the pre-training phase, as it enables the model to develop a more complete perception of visual information. By doing so, it enhances vision-language alignment and mitigates the information loss associated with relying solely on textual supervision.

\begin{table*}[th!]
\centering
\caption{\textbf{Extend experiments on LLaVA-Next dataset, LLaVA indicates the baseline (typically LVLM framework), ASVR builds upon the baseline by introducing autoregressive semantic visual supervision.}
"\ding{55}" indicates the use of textual supervision only. Visual encoder(SigLIP-ViT-SO400M/14@384 and SigLIP2-ViT-SO400M/14@384) are both utilized for ASVR and baseline to get different visual input and visual supervision.}
\tablestyle{5pt}{1.5}
\setlength\tabcolsep{2.2pt}
\resizebox{1\textwidth}{!}{
\begin{tabular}{c|c|c|c|c|cccc|cccc|cc|cc|cc|c}
\toprule
{\color[HTML]{333333} \textbf{}} & {\color[HTML]{333333} \textbf{Visual Encoder}} & {\color[HTML]{333333} \textbf{Visual Supervision}} & {\color[HTML]{333333} \textbf{LLM Backbone}} & {\color[HTML]{333333} \textbf{Data}} & {\color[HTML]{333333} \textbf{TVQA}} & {\color[HTML]{333333} \textbf{DVQA}} & {\color[HTML]{333333} \textbf{OCRB}} & {\color[HTML]{333333} \textbf{CQA}}  & {\color[HTML]{333333} \textbf{MMB}}  & {\color[HTML]{333333} \textbf{MME}}  & {\color[HTML]{333333} \textbf{SEED}} & {\color[HTML]{333333} \textbf{GQA}}  & {\color[HTML]{333333} \textbf{MMMU}} & {\color[HTML]{333333} \textbf{AI2D}} & {\color[HTML]{333333} \textbf{RQA}}  & {\color[HTML]{333333} \textbf{MMVP}} & {\color[HTML]{333333} \textbf{Hbench}} & {\color[HTML]{333333} \textbf{POPE}} & {\color[HTML]{333333} \textbf{AVG}}  \\ \cline{1-20}
{\color[HTML]{333333} LLaVA}     & {\color[HTML]{333333} Siglip-so400m-384}       & {\color[HTML]{333333} \ding{55}}                           & {\color[HTML]{333333} Vicuna-v1.5–7B}        & {\color[HTML]{333333} LLaVA–Next}    & {\color[HTML]{333333} 57.7}          & {\color[HTML]{333333} 40.7}          & {\color[HTML]{333333} 37.9}          & {\color[HTML]{333333} 42.6}          & {\color[HTML]{333333} 67.4}          & {\color[HTML]{333333} 71.5}          & {\color[HTML]{333333} 67.2}          & {\color[HTML]{333333} 61.8}          & {\color[HTML]{333333} 34.3}          & {\color[HTML]{333333} 65.3}          & {\color[HTML]{333333} 54.6}          & {\color[HTML]{333333} 32.8}          & {\color[HTML]{333333} 33.1}            & {\color[HTML]{333333} 86.4}          & {\color[HTML]{333333} 53.8}          \\
\rowcolor[HTML]{E1EAFF}
{\color[HTML]{333333} \textbf{ASVR}}      & {\color[HTML]{333333} Siglip-so400m-384}       & {\color[HTML]{333333} Semantic Siglip}             & {\color[HTML]{333333} Vicuna-v1.5–7B}        & {\color[HTML]{333333} LLaVA–Next}    & {\color[HTML]{333333} 58.6}          & {\color[HTML]{333333} 40.6}          & {\color[HTML]{333333} 39.7}          & {\color[HTML]{333333} 43.4}          & {\color[HTML]{333333} 67.9}          & {\color[HTML]{333333} 73.0}            & {\color[HTML]{333333} 67.5}          & {\color[HTML]{333333} 62.9}          & {\color[HTML]{333333} 34.2}          & {\color[HTML]{333333} 65.8}          & {\color[HTML]{333333} 55.4}          & {\color[HTML]{333333} 36.8}          & {\color[HTML]{333333} \textbf{39.2}}   & {\color[HTML]{333333} 85.9}          & {\color[HTML]{333333} 55.1}          \\
{\color[HTML]{333333} LLaVA}     & {\color[HTML]{333333} Siglip2-so400m-384}      & {\color[HTML]{333333} \ding{55}}                           & {\color[HTML]{333333} Vicuna-v1.5–7B}        & {\color[HTML]{333333} LLaVA–Next}    & {\color[HTML]{333333} 59.2}          & {\color[HTML]{333333} 41.8}          & {\color[HTML]{333333} 40.5}          & {\color[HTML]{333333} 46.3}          & {\color[HTML]{333333} 66.9}          & {\color[HTML]{333333} 74.0}            & {\color[HTML]{333333} 68.3}          & {\color[HTML]{333333} 62.7}          & {\color[HTML]{333333} 34.7}          & {\color[HTML]{333333} 66.4}          & {\color[HTML]{333333} \textbf{56.9}} & {\color[HTML]{333333} 33.3}          & {\color[HTML]{333333} 35.1}            & {\color[HTML]{333333} 86.1}          & {\color[HTML]{333333} 55.2}          \\
\rowcolor[HTML]{E1EAFF}
{\color[HTML]{333333} \textbf{ASVR}}      & {\color[HTML]{333333} Siglip2-so400m-384}      & {\color[HTML]{333333} Semantic Siglip-2}           & {\color[HTML]{333333} Vicuna-v1.5–7B}        & {\color[HTML]{333333} LLaVA–Next}    & {\color[HTML]{333333} \textbf{61.0}}   & {\color[HTML]{333333} \textbf{43.7}} & {\color[HTML]{333333} \textbf{44.8}} & {\color[HTML]{333333} \textbf{49.9}} & {\color[HTML]{333333} \textbf{70.2}} & {\color[HTML]{333333} \textbf{76.8}} & {\color[HTML]{333333} \textbf{69.5}} & {\color[HTML]{333333} \textbf{63.4}} & {\color[HTML]{333333} \textbf{36.3}} & {\color[HTML]{333333} \textbf{67.3}} & {\color[HTML]{333333} 56.7}          & {\color[HTML]{333333} \textbf{42.0}}   & {\color[HTML]{333333} 35.8}            & {\color[HTML]{333333} \textbf{86.8}} & {\color[HTML]{333333} \textbf{57.4}} \\ \midrule
\end{tabular}}
\label{siglip2}
\end{table*}

\paragraph{The Impact of visual supervision}
\label{app:supervision}
We further extend our experiments by employing discrete VQ-SigLIP2~\citep{tschannen2025siglip2multilingualvisionlanguage} as visual supervision, which provides richer and stronger semantic information, to verify that enhanced visual-semantic supervision can better scale the effectiveness of ASVR. To ensure fair comparison, we use LLaVA-Next~\citep{liu2024llava} as the training dataset under identical conditions, evaluating ASVR against the baseline with SigLIP~\citep{zhai2023siglip} as both visual input and supervision, as well as with SigLIP2~\citep{tschannen2025siglip2multilingualvisionlanguage} serving the same roles.

The results shown in Table~\ref{siglip2} clearly demonstrate that stronger visual semantic encoders lead to better performance when used for supervision. Specifically, ASVR with SigLIP-2 outperforms the baseline (LLaVA) with SigLIP-2 by an average of +2.2 points across 14 benchmarks. In comparison, ASVR with SigLIP improves over its baseline by +1.3 points. These results indicate that ASVR benefits more from stronger semantic supervision, and that pairing ASVR with more powerful semantic vision supervision further enhances its ability to improve visual understanding.

\end{document}